%% file: main.tex
\documentclass[journal]{IEEEtran}
\usepackage{ifpdf}
\usepackage{enumitem}

\usepackage{booktabs}
\usepackage{multicol}
\usepackage{multirow}
\usepackage{arydshln}
\usepackage{caption}
\usepackage{cite}

\ifCLASSINFOpdf
  \usepackage[pdftex]{graphicx}
\else
\fi
\usepackage{amsmath}
\usepackage{algorithmic}
\newcommand{\secref}[1]{\S\ref{#1}}
\usepackage[linesnumbered,ruled]{algorithm2e}
\usepackage{hyperref}
\input{preamble}

\begin{document}
%
\title{HYDRO: Towards Non-Reversible Face De-Identification Using a High-Fidelity Hybrid Diffusion and Target-Oriented Approach}
%
%
%

\author{Felix~Rosberg,
        Vitomir~Štruc,
        Cristofer~Englund,
        Eren~Erdal~Aksoy,
        and~Fernando~Alonso-Fernandez}

%
%

\markboth{Preprint}%
{Rosberg \MakeLowercase{\textit{et al.}}: HYDRO: Towards Non-Reversible Face De-Identification Using a High-Fidelity Hybrid Diffusion and Target-Oriented Approach}
%




\maketitle

\begin{abstract}

Target-oriented face de-identification models aim to anonymize the identity of a target individual across different images or video frames, such that the target can no longer be reliably recognized, while maintaining key characteristics of the visual data.
Such models commonly leverage generative encoder-decoder architectures to manipulate facial appearances, enabling them to produce realistic high-fidelity de-identification results, while ensuring considerable attribute-retention capabilities.  However, target-oriented models also carry the risk of inadvertently preserving subtle identity cues,  making them (potentially) reversible and susceptible to reconstruction attacks.  To address this problem, we introduce in this paper a novel (robust) face de-identification approach, called \OURSp, that combines target-oriented models with a dedicated diffusion process specifically designed to destroy any imperceptible information that may allow learning to reverse the de-identification procedure.~\OURSp~first de-identifies the given face image, injects noise into the de-identification result to impede reconstruction, and then applies a diffusion-based recovery step to improve fidelity and minimize the impact of the noising process on the data characteristics.~To further improve image fidelity and better retain gaze directions,  a novel Eye Similarity Discriminator (\ESD) is also introduced and incorporated it into the training of \OURSp. 
Extensive quantitative and qualitative experiments on three diverse datasets demonstrate that \OURSp~exhibits state-of-the-art (SOTA) fidelity and attribute-retention capabilities, while being the only target-oriented method resilient against reconstruction attacks. 
In comparison to multiple SOTA competitors, \OURSp~reduces the success of reconstruction attacks by $85.7\%$ on average.
\end{abstract}

\begin{IEEEkeywords}
De-Identification, Diffusion, Reconstruction Attacks.
\end{IEEEkeywords}

%
\IEEEpeerreviewmaketitle

\section{Introduction}
%
%
%
%

\IEEEPARstart{T}{he} rise of privacy regulations, like the General Data Protection Regulation \cite{GDPR}, has increased the demand for methods that safeguard individual privacy, while preserving the utility of visual data. In this context, face de-identification has emerged as a vital research area within computer vision, aiming to obscure or alter identifiable facial features while preserving essential attributes such as gender, expression or pose, among others \cite{FIVA, G2Face, FIT, RIDDLE, FALCO, PICLAnony, DEEPPRIVACY, DEEPPRIVACY2, MYFACEMYCHOICE, CFANET, CIAGAN, LEARNINGTOANON, DISENTANGLEDANON, DiffusionAnon2,ANONNET}. De-identification methods enable the sharing and analysis of visual data in a privacy-preserving manner, which is crucial for a wide range of applications in surveillance, police investigations, or healthcare \cite{meden2021privacy}.

Existing de-identification techniques often rely on low-level operations, such as pixelation, blurring, inpainting or obfuscation of the facial region \cite{meden2021privacy, DEEPPRIVACY, DEEPPRIVACY2}, to anonymize facial identities.~Such techniques typically ensure strong privacy protection, but often struggle with visual realism, data fidelity, and attribute retentions, which consequently leads to significantly reduced data utility for a number of data-driven problems and domains.~In response to these challenges, so-called \textit{target-oriented} face de-identification models  \cite{FaceDancer, FIVA, G2Face, simswap, FIT, LIVEDEID, CFANET} have recently emerged as a promising solution for the de-identification task.~Target-oriented techniques operate directly on the input (target) image and anonymize identities by manipulating facial features and appearances.~Techniques from this group commonly utilize generative encoder-decoder architectures to facilitate the de-identification process and excel in producing realistic, high-fidelity de-identified images that preserve important facial attributes and visual semantics.~Although target-oriented techniques have become the dominant de-identification methodology, they carry a \textbf{considerable risk of} retaining subtle identity cues that can render them reversible and prone to \textbf{reconstruction attacks}.~In fact, it has been shown in \cite{FIVA} that the privacy protection of a range of conceptually diverse target-oriented methods, i.e., \cite{FIVA, FaceDancer, simswap, MYFACEMYCHOICE}, can easily be circumvented by training adversarial models to reverse the de-identification procedure.

To address this fundamental risk of target-oriented models, we present in this paper a novel robust face de-identification model, named \OURSp.~The model uniquely combines the strengths of target-oriented models with a carefully designed diffusion process aimed at improving resilience against reconstruction attacks.~In the first step, \OURSp~adopts an encoder-decoder model to de-identify the input image in a target-oriented manner.~Here, an effective strategy is utilized to drive the de-identified face identity away from the input face by conditioning the generation process on identity-encoder embeddings, as shown in Fig.~\ref{fig:overview}.~Next, a one-step forward diffusion process is designed to infuse noise into the de-identified faces and conceal any potentially preserved identity information that could facilitate reconstruction attacks.~Finally, a backward diffusion process is learned to improve image fidelity and minimize the impact of the forward noising step on important data characteristics.~\OURSp~further improves attribute retention by incorporating a novel Eye Similarity Discriminator (ESD).
ESD is integrated into the training process of HYDRO to enhance image fidelity around the eye region and preserve gaze directions, which are crucial for maintaining the naturalness and utility of the de-identified faces.

The proposed \OURSp~model is evaluated through rigorous experiments on three datasets: FaceForensic++ \cite{faceforensics++}, CelebA \cite{CelebaA}, and Labeled Faces in the Wild (LFW) \cite{LFW} and achieves\textbf{ state-of-the-art (SOTA) performance on all three datasets} with respect to: (1) de-identification, (2) image fidelity, (3) attribute retention, and (4) resilience against reconstruction attacks.~Notably, \OURSp~stands out as the only target-oriented method that exhibits robustness against reconstruction attempts. In comparative experiments against multiple SOTA baselines \cite{FIVA, MYFACEMYCHOICE, FIT, G2Face, simswap}, it \textbf{reduces the average success rate} of reconstruction attacks \textbf{by $\mathbf{85.7\%}$} when compared to the closest competitor.

\section{Related Work}

Face de-identification methods can be broadly categorized into four main groups: inpainting-based \cite{CIAGAN, DEEPPRIVACY, DEEPPRIVACY2},  GAN-inversion-based \cite{DE-ID-VITOMIR, RIDDLE}, target-oriented \cite{FIVA, MYFACEMYCHOICE, CFANET, LIVEDEID, G2Face, FIT, DISENTANGLEDANON}, and diffusion-based \cite{DiffusionAnon1, DiffusionAnon2, DiffusionAnon3} methods.

\vspace{1.5mm}\noindent\textbf{Inpainting-based methods} typically first remove faces from images and then use generative models to fill in the resulting gap.~Since no identity information is ever revealed to the de-identification models, these methods provide strong privacy protection, but struggle to preserve facial attributes, as only contextual information is utilized for the inpainting process.~While attribute conditioning may help to mitigate this issue, this often increases processing time, as auxiliary models are required to extract the required (conditional) attribute information. A notable work from this group is DeepPrivacy \cite{DEEPPRIVACY, DEEPPRIVACY2}, which first removes the face from the input image and then generates a new one using a model conditioned on five key points.~This approach maintains the pose to some extent but neglects expression, gaze, gender, age, and other semantic details. Similarly, CIAGAN \cite{CIAGAN} faces comparable constraints, while also being confined to a fixed set of identities.

\vspace{1.5mm}\noindent\textbf{GAN-inversion-based methods} rely on a pre-trained Generative Adversarial Networks (GANs), such as StyleGAN \cite{stylegan, stylegan2, stylegan3}, for de-identifing faces in the GAN latent space. While  producing realistic and high-fidelity results, several notable limitations exist with these methods due to the required projection of images into the latent space. GAN-based models commonly misjudge occlusion due to glasses and hats, and struggle to identify and preserve attributes like gender, ethnicity, and even expression \cite{DE-ID-VITOMIR, RIDDLE}.
Meden \textit{et al.}~\cite{DE-ID-VITOMIR} addressed some of the aforementioned issues, but required numerous iterations of (time-consuming) latent optimizations to generate de-identified identities. 
RiDDLE \cite{RIDDLE} proposed an alternative optimization-free GAN-inversion-based method that relies on encryption in the latent space to ensure anonymity, but still inherits challenges due to the inversion process.

\vspace{1.5mm}\noindent\textbf{Target-oriented methods} commonly rely on encoder-decoder architectures for image manipulation, excelling in attribute retention and occlusion handling, while also exhibiting consistent performance on video data \cite{FIVA, FaceDancer, MYFACEMYCHOICE, LIVEDEID}.~These methods encompass face-swapping models, which encounter difficulties in gender and ethnicity retention if the conditional information is not correctly sampled~\cite{MYFACEMYCHOICE}, though counterfactual training can help \cite{FIVA, LIVEDEID}.~A key drawback of this group of methods is their \text{\em vulnerability to reconstruction attacks}, as detailed in \cite{FIVA}. Many SOTA methods, including FIVA \cite{FIVA}, G2Face \cite{G2Face}, Gafni \textit{et al.}~\cite{LIVEDEID}, and CFANET \cite{CFANET},  fall into this category and rely on some form of conditioning to guide the de-identification process towards a specific synthetic identity.

\vspace{1.5mm}\noindent\textbf{Diffusion-based methods} utilize the diffusion paradigm \cite{DDPM} to de-identify faces.~These methods can be considered specific instances of inpainting techniques, but typically exhibiting greater flexibility, while being constrained by the requirement for multiple sampling steps. 
Piano \textit{et al.}~\cite{DiffusionAnon1}, for instance, introduced two de-identification approaches based on Stable Diffusion \cite{StableDiffusion}, which, in turn, rely on several other auxiliary models, including the T2I-adapter \cite{T2IAdapter}, ControlNet \cite{ControlNet}, and the IP-adapter \cite{IPAdapter}.~Diff-Privacy \cite{DiffusionAnon3} proposed a reversible diffusion-based de-identification, while 
LDFA~\cite{DiffusionAnon2} introduced a pre-trained Stable Diffusion model without prompts for inpainting faces.
This latter approach effectively transforms Stable Diffusion into a computationally expensive inpainting model with limited attribute retention capabilities.

\vspace{1.5mm}\noindent\textbf{Gaze Retention} remains relatively understudied in the context of face de-identification. Prior research has addressed gaze retention in the context of face swapping \cite{EyeSim1UncannyPaper, GHOST}. GHOST \cite{GHOST}, for example, improved gaze retention by introducing a heatmap-based landmark detector \cite{AdaptiveWing} as a loss function. Wilson \textit{et al.} \cite{EyeSim1UncannyPaper}, on the other hand, incorporated an auxiliary gaze estimation model \cite{L2CSNetEyeGaze} into the training procedure of their face swapping model to facilitate better gaze preservation. Note that in certain applications, such as traffic safety, gaze retention (also in de-identified data) is essential for downstream tasks, as demonstrated by numerous prior studies \cite{GazeImportance1, GazeImportance2, GazeImportance3}, making it a key characteristic to preserve in during de-identification.

\section{Method}
\label{s:method}


In this section, we now present the main contribution of this work, i.e., \textbf{\OURSp, our novel target-oriented face de-identification approach} that offers distinct advantages over existing competitors in terms of attribute retention, visual semantics preservation, fidelity of the generated results and resilience against reconstruction attacks, as demonstrated later through rigorous experiments on multiple public datasets. A novel Eye Similarity Discriminator (\ESD) is also introduced to enhance gaze retention and contribute to the visual quality around the eye region of the de-identified faces.

\subsection{Overview of HYDRO}
\label{ss:netarchitecture}
\begin{figure}[!t]
\centering
\includegraphics[width=0.475\textwidth]{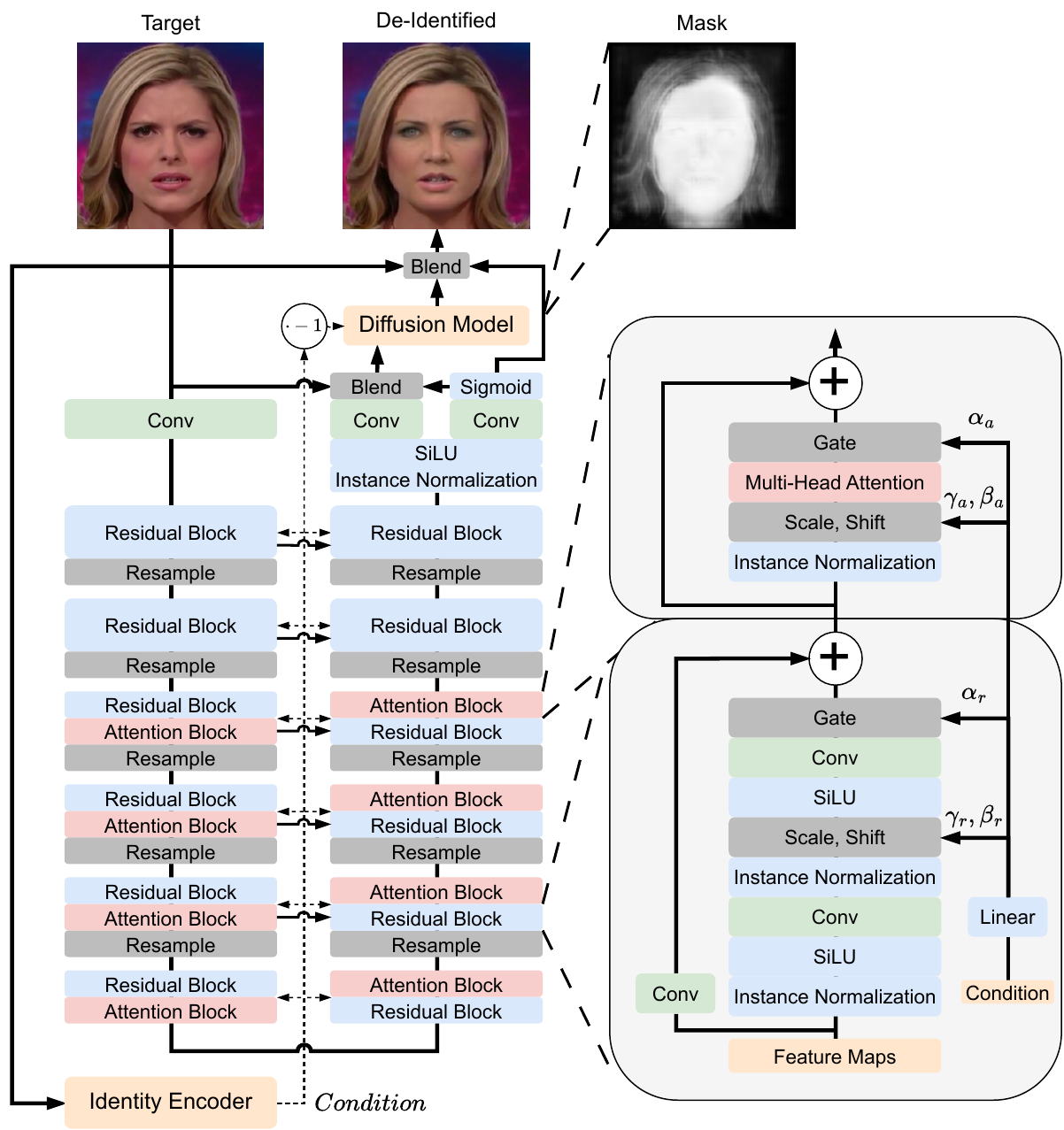}
\caption{\textbf{Overview of \OURSp.} \OURSp~relies on a U-Net architecture to de-identify the target image and produce a blending masks that is used to merge elements from the target with the de-identification result, thus promoting attribute preservation. A noising process is then applied to destroy potential identity traces that could facilitate reconstruction attacks. Finally, a diffusion-based denoising process is adopted to remove noise and improve image fidelity.
}
\label{fig:overview}
\end{figure}

\OURSp, illustrated in Fig.~\ref{fig:overview}, consists of three main components:~$(i)$ a target-oriented generator $G$ (i.e., the de-identification model) that takes a target face image as input and produces an initial de-identification result (\secref{sec:Generator}), $(ii)$ a diffusion model $\epsilon_{\theta}$ that impedes any potentially leaked identity information and improves image fidelity after de-identification (\secref{ss:method:reconstruction_robust}), and $(iii)$ an identity encoder $I$ that provides conditional identity information to control and learn the de-identification task (\secref{s:losses}).~To train HYDRO, two discriminators are adopted: a global residual discriminator and the proposed new eye similarity discriminator (\secref{ss:method:eye_similarity}).

\subsection{Target-Oriented Face De-Identification\label{sec:Generator}}

In the first step of the de-identification process, \OURSp~generates an initial version of the de-identified face using the target-oriented generator $G$.~Given a target face $x'$ to be de-identified  and an identity encoder $I$ (a face recognition embedding network), the generator produces two distinct outputs, i.e., a de-identified image $x_0'$ and a blending mask $m$, i.e., $G: x' \mapsto (x_0',m)$. The blending mask is used to combine the input face $x'$ with the generated image $x_0'$ and to ensure that: $(i)$ key visual attributes and semantics are propagated into the initial de-identification result $x_0$, $(ii)$ the matching background information is present in the de-identified images, and $(iii)$ the quality of the generated images is not compromised due to the presence of occlusions. Here, the blending process is defined as:
\begin{equation}
    x_0 = m \odot x_0' + (1-m)\odot x',
\end{equation}
where $\odot$ denotes the Hadamard product.~It is important to note that the blending mask $m$ is automatically learned  during the training process, jointly with the de-identification task and in an end-to-end manner, as detailed in Section~\secref{s:losses}. Thus, the blending procedure is an integral (and internal) part of the \OURS's design.

\vspace{1.5mm}\noindent\textbf{Generator Design:} To facilitate the de-identification procedure, we design the generator $G$ around a U-Net architecture with two output heads (one for $x_0'$ and one for $m$) that consists of a set of residual and attention blocks, each conditioned on identity information extracted from the identity encoder $I(x')$, as shown in Fig.~\ref{fig:overview}. As can be seen, the conditioning is implemented through a zero-initialized \cite{DiT} affine transform, which is used for scale and shift operations (governed by $\gamma_n$ and $\beta_n$ in Fig.~\ref{fig:overview}, respectively) \cite{styletransfer, stylegan, RectifiedFlowTransformerDiffusion, DiT} on the feature maps along with a gating operation (defined by $\alpha_n$) in both residual 
and attention blocks 
, where $n\in\{r,a\}$, similarly to \cite{RectifiedFlowTransformerDiffusion, DiT}. 
Here, zero-initialization is used to allow the residual blocks to begin as identity functions, while later gradually learning to inject identity information, which has been shown \cite{DiT, ScaleShiftGate} to allow for better control over the generated identities, enhance fidelity and reduce training time.

\subsection{Reconstruction Robustness through Diffusion \label{ss:method:reconstruction_robust}}

\noindent\textbf{Masking Identity Traces:}~The initial de-identified image $x_0$, produced by the generator $G$, already obscures the original identity and preserves important facial attributes (e.g., gender, ethnicity, and expression), while appearing realistic and visually convincing, as seen in Fig.~\ref{fig:overview}. However, as noted above and also emphasized in \cite{FIVA}, the generation process runs the risk of leaving identifiable traces in the synthesized image that can be discerned and exploited by adversarial models to circumvent the privacy protection. To address this issue, we destroy any remaining identity traces from the generator output by adding a small amount of noise according to the diffusion noising equation:
\begin{equation}
x_t=\alpha_tx_0+\sigma_t\epsilon_0,    
\label{eq:dif_noise}
\end{equation}
where $\epsilon_0\sim\mathcal{N}(0, I)$ represents noise sampled from a Gaussian distribution, $\alpha_t=\sqrt{\overline{\alpha}_{t}}$ and $\sigma_t=\sqrt{1-\overline{\alpha}_{t}} / \overline{\alpha}_{t}$ define the applied noise schedule at time step $t$, and $\overline{\alpha}_{t}$ is a cumulative product: $\overline{\alpha}_{t}=\prod_{i=0}^t(1-\beta_i)$, where $\beta_t$ is a variance parameter that controls how much noise is added at time instance $t$. To take advantage of the capabilities of diffusion models, while also minimizing the required sampling time, we implement the forward noising diffusion process using a single sampling step with low-strength noise, which was found empirically to be sufficient to minimize information loss and lead to competitive performance, when compared against the state-of-the-art. 

\vspace{1.5mm}\noindent\textbf{Diffusion-based Recovery:}~The injection of noise helps to impede reconstruction attacks, but adversely impacts the  image fidelity and low-level characteristics of the de-identified images. To mitigate this effect, we learn a backward diffusion process with the goal of improving image fidelity and recovering a realistic high-quality de-identified image.
Following the DPM-Solver++ \cite{DPMSolver++} equation, let $h$ denote $h=log(\alpha_{t-1}) - log(\sigma_{t-1})-log(\alpha_{t})+log(\sigma_{t})$, let $x_t=\alpha_tx_0+\sigma_t\epsilon_0$ again represent the noisy sample of $x_0$, subject to the noise schedule defined by $\alpha_t$ and $\sigma_t$, and let $\epsilon_\theta(x_t, t)$ denote the utilized diffusion model, parameterized by $\theta$. The denoising equation (backward diffusion process) for the DPM-Solver++ can then be written as:
\begin{equation}
  x_{t-1} = \frac{\sigma_{t-1}}{\sigma_{t}}x_t - \alpha_{t-1}(e^{-h}-1)\frac{x_t - \sigma_t\epsilon_{\theta}(x_t, t)}{\alpha_t}~.
  \label{eq:dpmsolver}
\end{equation}

We specifically look at the special case of the final step in the reverse diffusion schedule, where $\sigma_{t-1}=0$ and $\alpha_{t-1}=1$, thus, computationally $h \rightarrow\infty$, in the final stage:
\begin{equation}
  x_{t-1} = x_0-\frac{\sigma_t\epsilon_{\theta}(x_t, t)-\sigma_t\epsilon_0}{\alpha_t}~,
  \label{eq:special_dpmsolver}
\end{equation}
where, in the last step, $\hat{x}_0 = x_{t-1}$ is the image recovered from $x_t$. Finally, $\hat{x}_0$ is blended with the input $x'$,  similarly as described in Section \secref{sec:Generator} and shown in Fig.~\ref{fig:overview}.

\vspace{1.5mm}\noindent\textbf{Diffusion-Model Design:}~The diffusion model, $\epsilon_{\theta}$, is implemented with a U-Net architecture similar to the generator $G$ and is also conditioned on the identity information provided by the identity encoder $I$. A distinction here lies in the way the conditional information is applied. Since the main goal of the diffusion model in the backward process is to recover a high-quality denoised image while preventing reconstruction attacks, the identity condition is reversed in the condition hyper-sphere by multiplying the embedding vector by $-1$ during inference to avoid running the identity encoder twice and avoid risks of injecting the true identity.

\vspace{1.5mm}\noindent\textbf{Privacy-Protection Strength:}~Assuming the worst-case scenario, where the noise schedule of $\alpha$ and $\sigma$ are known, it is clear that a significant amount of information about the diffusion process is still required to recover an exact version of the initially de-identified image $x_0$ that would allow for a successful reconstruction attack. By rearranging Eq.~\eqref{eq:special_dpmsolver}:
\begin{equation}
  x_0 = x_{t-1}+\frac{\sigma_t\epsilon_{\theta}(x_t, t)-\sigma_t\epsilon_0}{\alpha_t}~,
  \label{eq:special_reverse_dpmsolver}
\end{equation}
it is evident that access to both the exact applied noise $\epsilon_0$ and the predicted noise from the diffusion model $\epsilon_{\theta}(x_t, t)$, are required for the prediction of $x_0$.~Therefore, the precise recovery of $x_0$ is unfeasible in the absence of access to $x_t$, $\epsilon_0$, and the diffusion model, which provides a strong foundation for the  privacy protection ensured by \OURSp.

\subsection{Eye Similarity Discriminator (ESD)}
\label{ss:method:eye_similarity}
\begin{figure}[!t]
\centering
\includegraphics[width=0.99\columnwidth,trim = 14mm 2mm 17mm 0, clip]{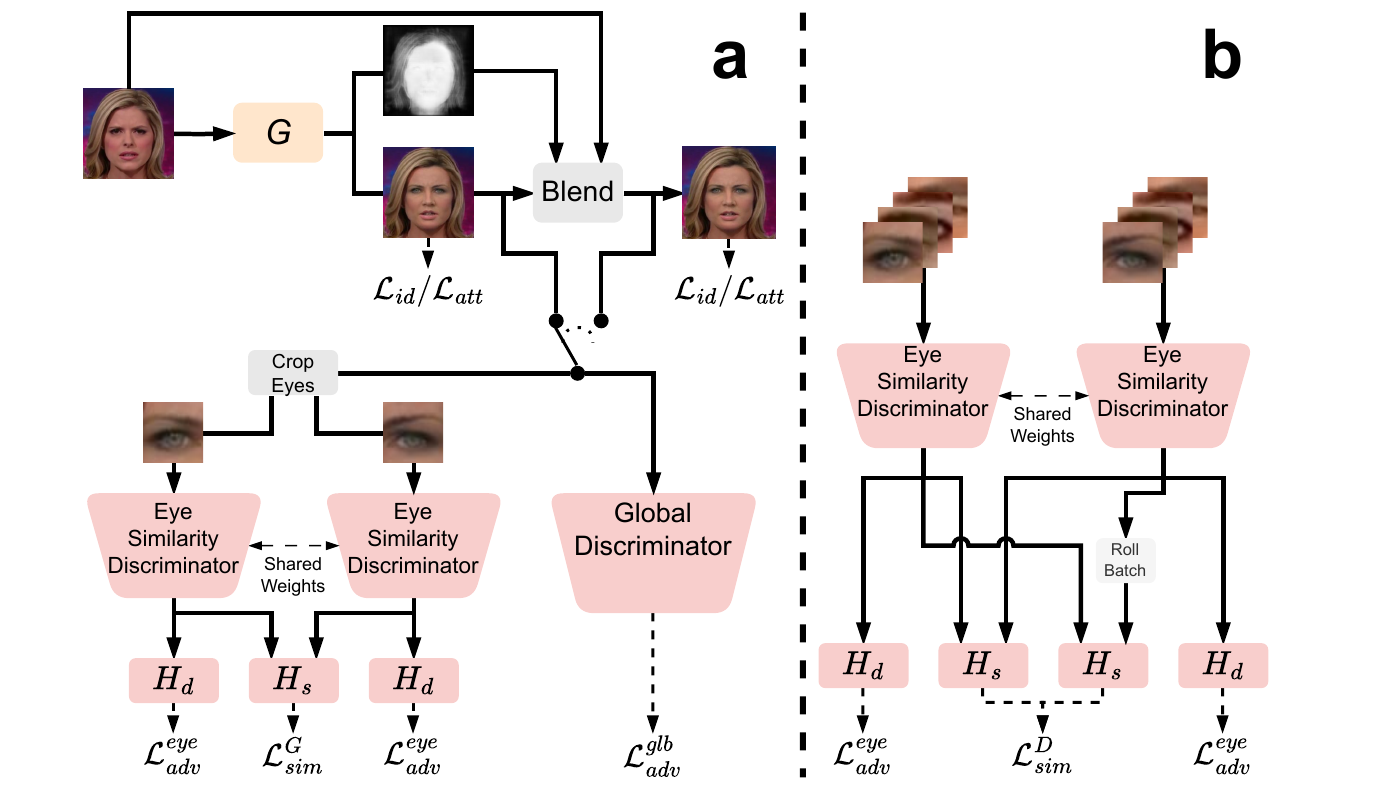}
\caption{\textbf{Training of the (a) target-oriented generator $G$ and (b) the \ESD.} Note that (a) is for fake samples, whereas (b) processes real samples. $H_d$ and $H_s$ are the discrimination and similarity heads.\vspace{-1mm}}
\label{fig:training}
\end{figure}

The quality of the reconstructed eyes and the retention of the eye gaze play an important role in preserving the visual quality and attributes of the original input while de-identifying faces.
To this end, we introduce an additional  discriminator that focuses specifically on the crops of the left and right eyes. 
The discriminator comprises a backbone, a discrimination head $H_d$, and a similarity head $H_s$. During training, the eye location is detected and extracted into $32\times 32$ patches using the face alignment procedure from \cite{AdaptiveWing}. The feature maps from the backbone are passed to the discriminator head, producing logits for both the left and right eyes. The model learns to distinguish between the same and different eyes in a contrastive manner by exploiting the left and right eye as a positive pair and a rolled batch of both eyes as negative pairs, as shown in Fig. \ref{fig:training}b. The similarity head takes concatenated features of both eyes as extracted by the backbone to produce the similarity logits. We use binary cross-entropy to train the discriminator to score similarity for both eyes as follows:
\begin{equation}
\resizebox{0.9\hsize}{!}{$
  \mathcal{L}^{D}_{sim} = bce(H_s(cat(f_{l}, f_{r})), 1) 
  + bce(H_s(cat(f_{l}, roll(f_{r}))), 0)~,$
  \label{eq:eye_sim_loss_d}
}
\end{equation}
where $bce$ represents the binary cross-entropy, $cat$ denotes concatenation, $roll$ is the operation for rolling the batch in the batch dimension, $f_l$ and $f_r$ are the feature maps of the left and right eyes, respectively. The similarity is employed to train the generator $G$ to further improve the visual quality of the eye region and by extension the gaze retention as:
\begin{equation}
\begin{split}
  \mathcal{L}^{G}_{sim} = bce(H_s(cat(f_{l}, f_{r})), 1)~,
  \label{eq:eye_sim_loss_g}
\end{split}
\end{equation}
where $f_l$ and $f_r$ are from the output of $G$ (See Fig. \ref{fig:training}a).

\subsection{Loss Functions and Training Procedure}
\label{s:losses}

\noindent\textbf{Generator Losses:}~To train the target-oriented generator $G$, we utilize:~$(i)$ an identity loss ($\mathcal{L}_{id}$) for driving the identity away from the input face image $x'$, $(ii)$ an attribute-preservation loss ($\mathcal{L}_{att}$) for retaining facial attributes, $(iii)$ an adversarial loss ($\mathcal{L}_{adv}$) for fidelity, and $(iv)$ an eye similarity loss ($\mathcal{L}^{G}_{sim}$) for improving gaze retention, eye fidelity and eye symmetry, as shown in Fig.~\ref{fig:training}a.~We apply all loss functions on the initial de-identified image $x_0'$ and the blended result $x_0$ (see Fig. \ref{fig:training}a), which allows us to implicitly learn the blending mask $m$ without explicit supervision. To prevent the  mask from collapsing during training, 
the discriminator losses are interleaved between the generator output $x_0'$ and the blended result $x_0$ every other batch.

The following losses are employed to train \OURS:
\begin{itemize}[noitemsep,leftmargin=*]
    \item With the \textbf{identity loss} $\mathcal{L}_{id}$, we aim to minimize the cosine similarity between the input and output of $G$ as:

\begin{equation}
  \mathcal{L}_{id} = 1 + cos(I(x'), I(x^*))~,
  \label{eq:identity_loss}
\end{equation}

where $x'$ is the input face, $x^*\in\{x_0, x_0'\}$ is the de-identified face and $I(.)$ is the identity encoder in the form of a ResNet-$100$ model trained with the ArcFace loss on the MS1MV3 dataset \cite{arcface}. $cos(.)$ denotes the cosine similarity between the two identity vectors $I(x')$ and $I(x^*)$. 

\item For the \textbf{attribute-preservation loss} $\mathcal{L}_{att}$, we use a combination of different losses that try to encourage the generator to retain as much of the attribute information from the input image $x'$ in the de-identified results ($x_0$ and $x_0'$) as possible. First, we adopt a \textit{feature-similarity loss} $\mathcal{L}_{fsr}$ that aims to minimize the cosine distance between the intermediate feature representations of $x'$ and $x^*\in\{x_0,x_0'\}$, extracted by the ArcFace model \cite{FIVA, FaceDancer}. Next, we employ an L1 \textit{reconstruction loss} $\mathcal{L}_{rec}$ between the input image $x'$ and de-identified outputs $x^*\in\{x_0,x_0'\}$ similarly to \cite{FIVA, LIVEDEID} and apply an L1 \textit{regularization term} $\mathcal{L}_{msk}$ over the generated mask to control pixel magnitudes. To preserve specific attributes, we use an L1 \textit{distance loss} $\mathcal{L}_{shp}$ between expression and pose coefficients extracted from a 3DMM encoder \cite{3DMMEncoder}. Finally, we also incorporate an L1 \textit{distance gaze loss} $\mathcal{L}_{gze}$ between left and right eye heatmaps extracted by a heatmap-based landmark encoder \cite{AdaptiveWing}. The resulting attribute loss $\mathcal{L}_{att}$ is:

\begin{equation}
\resizebox{.87\hsize}{!}{$
  \mathcal{L}_{att} = \mathcal{L}_{fsr} + \lambda_{rec}\mathcal{L}_{rec} + \lambda_{msk}\mathcal{L}_{msk} \\
  + \lambda_{shp}\mathcal{L}_{shp} + \lambda_{gze}\mathcal{L}_{gze} ~,$
  \label{eq:attribute_loss}
}
\end{equation}

where the balancing weights $\lambda_{rec}=0.5$, $\lambda_{msk}=0.003$, $\lambda_{shp}=1$, and $\lambda_{gze}=25$ were determined heuristically through preliminary experiments. 
\end{itemize}

\vspace{1.5mm}\noindent\textbf{Discriminators:} We rely on two discriminators, when training \OURSp. The first is a \textit{global discriminator} that is responsible for evaluating the realism of the entire image and has the same structure as the discriminators used in prior generative face-swapping  works \cite{FIVA, FaceDancer, hififace, stargan}. Details on the global discriminator are available in the supplementary material. The second is the proposed \textit{eye similarity discriminator} (\ESD). Following established practice \cite{FaceDancer, simswap}, we train the discriminators using hinge losses \cite{HingeLoss}. At each iteration, the global discriminator is fed with a batch of real samples and a batch of fake samples. The \ESD~is fed with a batch of real samples for both eyes, and a batch of fake samples for both eyes. \ESD~ is also trained contrastively to score left and right eyes. As seen in Fig. \ref{fig:training}b, we concatenate feature maps of both eyes and pass them to the similarity head to evaluate the final similarity. We exploit the left and right eyes as positive pairs, while for negative pairs we simply roll the batch in the batch dimension. For both discriminators, we also apply gradient penalty regularization $\mathcal{L}_{gp}$ with a loss weight of $\lambda_{gp}=0.005$.

\vspace{1.5mm}\noindent\textbf{Overall Training Objective:} The two discriminators provide adversarial (hinge) losses $\mathcal{L}_{adv}$ for the training of the de-deidentification model $G$, while ESD also defines the eye-similarity loss $\mathcal{L}^{G}_{sim}$ (in Eq.~\eqref{eq:eye_sim_loss_g}) that jointly with the generator losses define the learning objective of $G$, i.e.:   
\begin{equation}
\resizebox{.88\hsize}{!}{$
  \mathcal{L} = \mathcal{L}_{att} + \lambda_{id}\mathcal{L}_{id} +  \lambda^{glb}_{adv}\mathcal{L}^{glb}_{adv} + \lambda^{eye}_{adv}\mathcal{L}^{eye}_{adv} \\
  + \lambda^{G}_{sim}\mathcal{L}^{G}_{sim}~,$
  \label{eq:total_loss}
}
\end{equation}
with $\lambda_{id}=5$, $\lambda^{glb}_{adv}=1$, $\lambda^{eye}_{adv}=0.5$, and $\lambda^{G}_{sim}=0.5$.

\section{Reconstruction-Attack Model}
\label{sec:attack model}

\begin{figure}[!t]
\centering
\includegraphics[width=0.45\textwidth, trim = 0 0 0 0mm, clip]{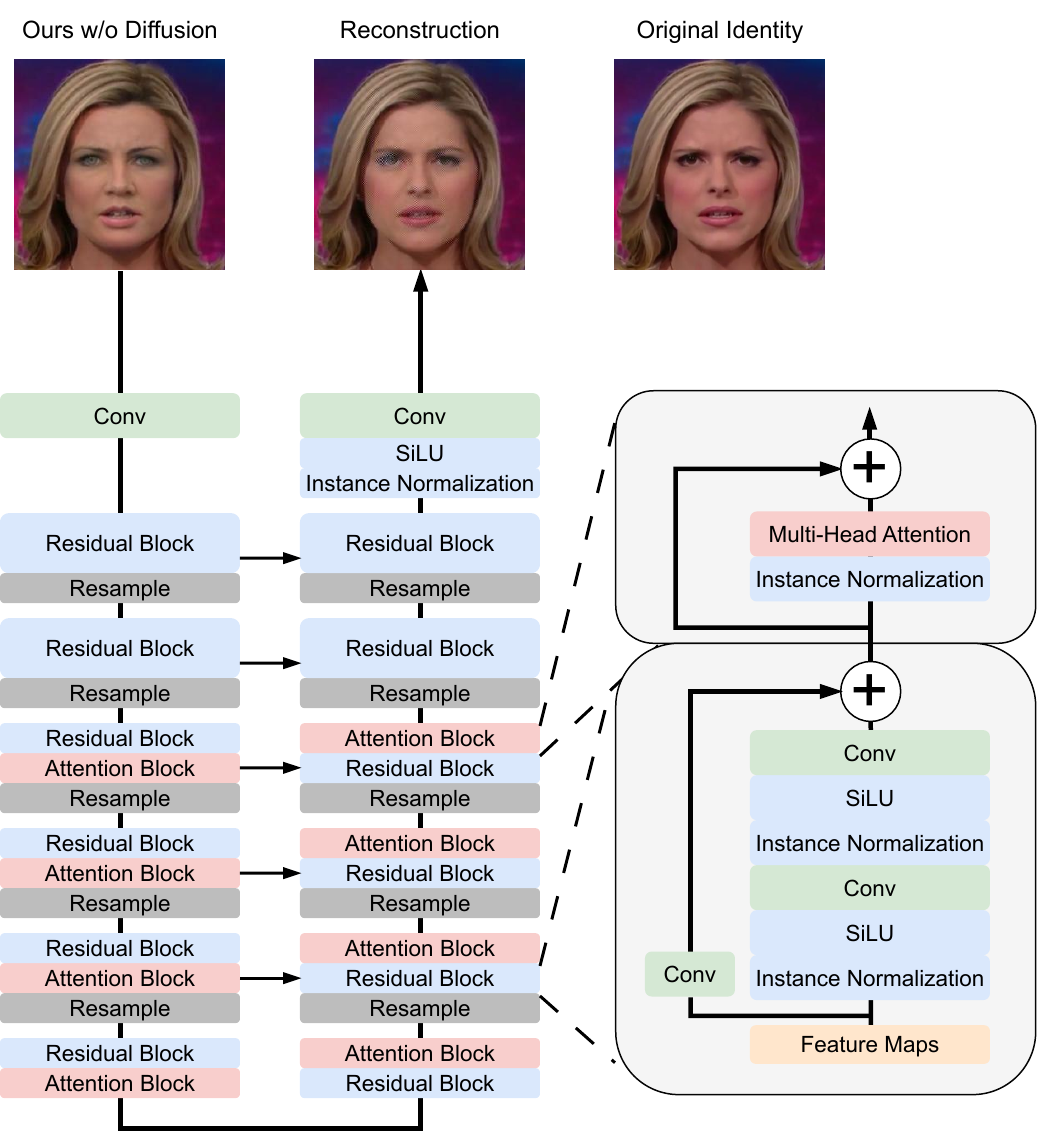}
\caption{\textbf{Overview of the reconstruction-attack model $R$.} In this particular case, we illustrate HYDRO \emph{without} the diffusion process, which results in a reconstruction that causes a match using face recognition models.
}
\label{fig:reconstructor}
\end{figure}

A key component of the evaluation of HYDRO is the Reconstruction-Attack Model $R$ that is used to probe the robustness of HYDRO and other competing de-identification models to reconstruction attempts.~Below, we discuss the architectural design of the model, as well as  the training procedure used to learn its parameters.  

\begin{table*}[!t]
\caption{\textbf{Quantitative results} on FaceForensic++~\cite{faceforensics++}, CelebA~\cite{CelebaA}, and  LFW~\cite{LFW}. \textbf{Bold} indicates best, and \underline{underline} shows second best results. \vspace{-2mm}
}
\begin{center}
\resizebox{\textwidth}{!}{
\begin{tabular}{ll|ccccccc:ccccccccccccccc:ccc}
\toprule
&\multirow{3}{*}{Model} &&&&&&&&&&&&&&&                                                                                                                                        \multicolumn{4}{c}{Identity Retrieval$\downarrow$} \\
&& FID$\downarrow$ & Pose$\downarrow$ & Pose$\downarrow$ & Exp$\downarrow$ & Gaze$\downarrow$ & MSE$\downarrow$ & GFLOPS$\downarrow$ & \multicolumn{3}{c}{CosFace} & \multicolumn{3}{c}{ArcFace} & \multicolumn{3}{c}{FaceNet} & \multicolumn{3}{c}{AdaFace} & \multicolumn{3}{c}{ElasticFace} & \multicolumn{3}{c}{Average} \\

&            & & \cite{3DMMEncoder} & \cite{HOPENET} & \cite{3DMMEncoder} & \cite{AdaptiveWing} & & & \tiny $10^{-3}$ & \tiny $10^{-4}$ & \tiny $10^{-5}$ & \tiny $10^{-3}$ & \tiny $10^{-4}$ & \tiny $10^{-5}$ & \tiny $10^{-3}$ & \tiny $10^{-4}$ & \tiny $10^{-5}$ & \tiny $10^{-3}$ & \tiny $10^{-4}$ & \tiny $10^{-5}$ & \tiny $10^{-3}$ & \tiny $10^{-4}$ & \tiny $10^{-5}$ & \tiny $10^{-3}$ & \tiny $10^{-4}$ & \tiny $10^{-5}$ \\

\cmidrule{2-27}
\parbox[t]{1mm}{\multirow{7}{*}{\rotatebox[origin=c]{90}{\normalsize FaceForensic++}}}
&DeepPrivacy \cite{DEEPPRIVACY} \small{(ISVC'19)}  & 30.77            & 0.05          & 3.52              & 3.33              & 13.38 & 0.05 & 122.78             & 0.08          & 0.05          & \textbf{0.00} & 0.06          & 0.03          & 0.01          & 0.04          & \textbf{0.00} & \textbf{0.00} & 0.06 & 0.04 & 0.02 & 0.07 & 0.05 & 0.02 & 0.06 & 0.04 & 0.01 \\
&CIAGAN \cite{CIAGAN} \small{(CVPR'20)}            & 25.75            & 0.06          & 3.39              & 2.76              & 14.73 & 0.28 & \textbf{3.30}             & \textbf{0.00} & \textbf{0.00} & \textbf{0.00} & \textbf{0.00} & \textbf{0.00} & \textbf{0.00} & \textbf{0.00} & \textbf{0.00} & \textbf{0.00} & \textbf{0.00} & \textbf{0.00} & \textbf{0.00} & \textbf{0.00} & \textbf{0.00} & \textbf{0.00} & \textbf{0.00} & \textbf{0.00} & \textbf{0.00} \\
&RiDDLE \cite{RIDDLE} \small{(CVPR'23)}            & 37.81            & 0.06          & 3.52              & 3.43              & 12.00 & 0.19 & 301.01             & \textbf{0.00} & \textbf{0.00} & \textbf{0.00} & \textbf{0.00} & \textbf{0.00} & \textbf{0.00} & \textbf{0.00} & \textbf{0.00} & \textbf{0.00} & \textbf{0.00} & \textbf{0.00} & \textbf{0.00} & \textbf{0.00} & \textbf{0.00} & \textbf{0.00} & \textbf{0.00} & \textbf{0.00} & \textbf{0.00} \\
&G2Face \cite{G2Face} \small{(TIFS'24)}            & 9.41             & \textbf{0.03} & 1.94              & \underline{2.63}  & \underline{6.76} & 0.03 & 280.02  & \textbf{0.00} & \textbf{0.00} & \textbf{0.00} & \textbf{0.00} & \textbf{0.00} & \textbf{0.00} & 0.01          & \textbf{0.00} & \textbf{0.00} & \textbf{0.00} & \textbf{0.00} & \textbf{0.00} & \textbf{0.00} & \textbf{0.00} & \textbf{0.00} & \textbf{0.00} & \textbf{0.00} & \textbf{0.00} \\
&FIVA \cite{FIVA} \small{(ICCV'23)}                & \underline{4.29} & \textbf{0.03} & \underline{1.90}  & 2.84              & 7.24 & \textbf{0.02} & 304.23              & \textbf{0.00} & \textbf{0.00} & \textbf{0.00} & \textbf{0.00} & \textbf{0.00} & \textbf{0.00} & 0.01          & \textbf{0.00} & \textbf{0.00} & \textbf{0.00} & \textbf{0.00} & \textbf{0.00} & \textbf{0.00} & \textbf{0.00} & \textbf{0.00} & \textbf{0.00} & \textbf{0.00} & \textbf{0.00} \\
&FIT \cite{FIT} \small{(ECCV'20)}                  & 20.72            & 0.04          & 2.71              & 2.69              & 10.01 & 0.19 & \underline{30.09}             & 0.14          & 0.08          & 0.01          & 0.10          & 0.06          & 0.02          & 0.03          & \textbf{0.00} & \textbf{0.00} & 0.13 & 0.09 & 0.04 & 0.12 & 0.10 & 0.05 & 0.10 & 0.07 & 0.02 \\
&\OURSp~(Ours)                                     & \textbf{3.55}    & \textbf{0.03} & \textbf{1.77}     & \textbf{2.22}     & \textbf{6.00} & \textbf{0.02} & 654.19     & \textbf{0.00} & \textbf{0.00} & \textbf{0.00} & \textbf{0.00} & \textbf{0.00} & \textbf{0.00} & \textbf{0.00}& \textbf{0.00}& \textbf{0.00} & \textbf{0.00}& \textbf{0.00}& \textbf{0.00}& \textbf{0.00}& \textbf{0.00}& \textbf{0.00} & \textbf{0.00} & \textbf{0.00} & \textbf{0.00} \\
\cmidrule{2-27}
\parbox[t]{1mm}{\multirow{7}{*}{\rotatebox[origin=c]{90}{\normalsize CelebA}}}
&DeepPrivacy \cite{DEEPPRIVACY} \small{(ISVC'19)}  & 10.65            & 0.08              & 4.76              & 2.98              & 13.75 & 0.08 & 122.78             & 0.05          & 0.03          & \textbf{0.00} & 0.04          & 0.02          & 0.01          & 0.04          & \textbf{0.00} & \textbf{0.00} & 0.04 & 0.03 & 0.01 & 0.04 & 0.04 & 0.01 & 0.04 & 0.03 & 0.01 \\
&CIAGAN \cite{CIAGAN} \small{(CVPR'20)}            & 10.61            & 0.07              & 4.15              & 2.75              & 16.10 & 0.35 & \textbf{3.30}             & 0.07          & 0.05          & \textbf{0.00} & 0.04          & 0.02          & \textbf{0.00} & 0.02          & \textbf{0.00} & \textbf{0.00} & 0.05 & 0.03 & 0.01 & 0.05 & 0.04 & 0.01 & 0.04 & 0.03 & 0.00 \\
&RiDDLE \cite{RIDDLE} \small{(CVPR'23)}            & 38.22            & 0.08              & 4.98              & 3.25              & 13.27 & 0.21 & 301.01             & \textbf{0.00} & \textbf{0.00} & \textbf{0.00} & \textbf{0.00} & \textbf{0.00} & \textbf{0.00} & \textbf{0.00} & \textbf{0.00} & \textbf{0.00} & \textbf{0.00} & \textbf{0.00} & \textbf{0.00} & \textbf{0.00} & \textbf{0.00} & \textbf{0.00} & \textbf{0.00} & \textbf{0.00} & \textbf{0.00} \\
&G2Face \cite{G2Face} \small{(TIFS'24)}            & 4.90             & \underline{0.04}  & 2.66              & 2.59              & 7.32 & 0.03 & 280.02              & \textbf{0.00} & \textbf{0.00} & \textbf{0.00} & \textbf{0.00} & \textbf{0.00} & \textbf{0.00} & \textbf{0.01} & \textbf{0.00} & \textbf{0.00} & 0.01 & \textbf{0.00} & \textbf{0.00} & \textbf{0.00} & \textbf{0.00} & \textbf{0.00} & 0.01 & \textbf{0.00} & \textbf{0.00} \\
&FIVA \cite{FIVA} \small{(ICCV'23)}                & \underline{2.04} & \underline{0.04}  & \underline{2.40}  & 2.78              & \underline{7.24} & \textbf{0.02} & 304.23  & \textbf{0.00} & \textbf{0.00} & \textbf{0.00} & \textbf{0.00} & \textbf{0.00} & \textbf{0.00} & \textbf{0.01} & \textbf{0.00} & \textbf{0.00} & \textbf{0.00} & \textbf{0.00} & \textbf{0.00} & \textbf{0.00} & \textbf{0.00} & \textbf{0.00} & \textbf{0.00} & \textbf{0.00} & \textbf{0.00} \\
&FIT \cite{FIT} \small{(ECCV'20)}                  & 10.23            & \underline{0.04}  & 2.89              & \underline{2.48}  & 8.73 & 0.09 & \underline{30.09}              & 0.12          & 0.10          & 0.02          & 0.08          & 0.05          & 0.02          & 0.04          & \textbf{0.00} & \textbf{0.00} & 0.12 & 0.09 & 0.05 & 0.11 & 0.10 & 0.05 & 0.09 & 0.07 & 0.03 \\
&\OURSp~(Ours)                    & \textbf{1.14}    & \textbf{0.03}     & \textbf{2.19}     & \textbf{2.16}     & \textbf{6.13} & \textbf{0.02} & 654.19     & \textbf{0.00} & \textbf{0.00} & \textbf{0.00} & \textbf{0.00} & \textbf{0.00} & \textbf{0.00} & \textbf{0.00} & \textbf{0.00} & \textbf{0.00} & \textbf{0.00}& \textbf{0.00}& \textbf{0.00}& \textbf{0.00}& \textbf{0.00}& \textbf{0.00} & \textbf{0.00} & \textbf{0.00} & \textbf{0.00} \\
\cmidrule{2-27}
\parbox[t]{1mm}{\multirow{7}{*}{\rotatebox[origin=c]{90}{\normalsize LFW}}}
&DeepPrivacy \cite{DEEPPRIVACY} \small{(ISVC'19)}  & 9.33             & 0.06              & 4.31              & 3.08              & 13.47 & 0.06 & 122.78             & 0.07          & 0.02          & \textbf{0.00} & 0.06          & 0.02          & \textbf{0.00} & 0.04          & \textbf{0.00} & \textbf{0.00} & 0.06          & 0.02          & \textbf{0.00} & 0.07 & 0.03 & 0.01 & 0.06 & 0.02 & \textbf{0.00} \\
&CIAGAN \cite{CIAGAN} \small{(CVPR'20)}            & 20.55            & 0.08              & 5.89              & 3.25              & 29.85 & 0.42 & \textbf{3.30}             & 0.08          & 0.01          & \textbf{0.00} & 0.06          & \textbf{0.00} & \textbf{0.00} & 0.01          & \textbf{0.00} & \textbf{0.00} & 0.05          & \textbf{0.00} & \textbf{0.00} & 0.08 & 0.03 & \textbf{0.00} & 0.06 & 0.01 & \textbf{0.00} \\
&RiDDLE \cite{RIDDLE} \small{(CVPR'23)}            & 64.31            & 0.07              & 4.70              & 3.50              & 13.80 & 0.27 & 301.01             & \textbf{0.00} & \textbf{0.00} & \textbf{0.00} & \textbf{0.00} & \textbf{0.00} & \textbf{0.00} & \textbf{0.00} & \textbf{0.00} & \textbf{0.00} & \textbf{0.00} & \textbf{0.00} & \textbf{0.00} & \textbf{0.00} & \textbf{0.00} & \textbf{0.00} & \textbf{0.00} & \textbf{0.00} & \textbf{0.00} \\
&G2Face \cite{G2Face} \small{(TIFS'24)}            & 6.75             & \underline{0.04}  & 2.61              & 2.77              & \underline{7.59} & 0.03 & 280.02  & \textbf{0.00} & \textbf{0.00} & \textbf{0.00} & 0.01          & \textbf{0.00} & \textbf{0.00} & 0.01          & \textbf{0.00} & \textbf{0.00} & 0.01          & \textbf{0.00} & \textbf{0.00} & 0.01 & \textbf{0.00} & \textbf{0.00} & 0.01 & \textbf{0.00} & \textbf{0.00} \\
&FIVA \cite{FIVA} \small{(ICCV'23)}                & \underline{4.40} & \underline{0.04}  & \underline{2.59}  & 3.04              & 7.83 & \textbf{0.02} & 304.23              & \textbf{0.00} & \textbf{0.00} & \textbf{0.00} & \textbf{0.00} & \textbf{0.00} & \textbf{0.00} & \textbf{0.00} & \textbf{0.00} & \textbf{0.00} & \textbf{0.00} & \textbf{0.00} & \textbf{0.00} & \textbf{0.00} & \textbf{0.00} & \textbf{0.00} & \textbf{0.00} & \textbf{0.00} & \textbf{0.00} \\
&FIT \cite{FIT} \small{(ECCV'20)}                  & 9.40             & \underline{0.04}  & 3.09              & \underline{2.61}  & 9.26 & 0.10 & \underline{30.09}              & 0.21          & 0.09          & 0.01          & 0.16          & 0.07          & 0.02          & 0.02          & \textbf{0.00} & \textbf{0.00} & 0.17          & 0.08          & 0.03          & 0.16 & 0.09 & 0.04 & 0.14 & 0.06 & 0.02 \\
&\OURSp~(Ours)                    & \textbf{3.08}    & \textbf{0.03}     & \textbf{2.31}     & \textbf{2.28}     & \textbf{6.34} & \textbf{0.02} & 654.19     & \textbf{0.00} & \textbf{0.00} & \textbf{0.00} & \textbf{0.00} & \textbf{0.00} & \textbf{0.00} & 0.01          & \textbf{0.00} & \textbf{0.00} & \textbf{0.00} & \textbf{0.00} & \textbf{0.00} & \textbf{0.00}& \textbf{0.00}& \textbf{0.00} & \textbf{0.00} & \textbf{0.00} & \textbf{0.00} \\
\bottomrule 
\end{tabular}
}\vspace{-2mm}
\end{center}
\label{t:compare}
\end{table*}

\vspace{1.5mm}\noindent\textbf{Reconstruction Attack Model Architecture:} The reconstruction attack model $R$ is based on the U-Net model and shares the majority of its architectural details with the generator $G$ and diffusion model $\epsilon_{\theta}$ as shown in Fig. \ref{fig:reconstructor}. However, there are two key differences: $(i)$ first, the reconstruction-attack model is an unconditional U-Net and, hence, is not conditioned on any information (e.g., identity, time-steps, etc.) and $(ii)$ second, it contains a single head that is responsible for predicting the initial target image before de-identification that still contains traces of the original identity and is, therefore, susceptible to reconstruction attacks. Given that a single head is used, the model obviously also does not return a mask.

Let the initial target face image be $x$ and let the corresponding de-identification version be denoted as $x_{d}$. The goal of $R$ is then to learn a mapping that can recover an approximation of the target face image, i.e.: $R: x_{d} \mapsto \hat{x}$, where $\hat{x}$ is expected to contain the same identity information as $x$ in the case of a successful reconstruction attack.

\vspace{1.5mm}\noindent\textbf{Reconstruction Attack Model Training:} The reconstruction attack model is trained over pairs of target face images and the respective de-identified counterparts. The training procedure assumes a \textit{black-box setting} simulating a bad-actor getting access to a sizable number of input and output images.  
One reconstruction model is trained for each state-of-the-art model that we compared with in the experimental evaluation conducted in the main manuscript. When learning the attack model for HYDRO, we utilize de-identified samples, on which the diffusion step was already applied. This setting is based on the assumption that if there is any leaked identity information, the reconstruction attack model will reliably learn to recover the original target identity.

Each reconstruction-attack model is trained using a combination of three losses, i.e., an L1 reconstruction loss, a perceptual loss \cite{LPIPS}, and an identity loss, similar to the one in the main manuscript, Eq.~8. The role of each loss is defined as follows:

\begin{itemize}
\item The L1 \textbf{reconstruction loss} $\mathcal{L}^R_{rec}$ encourages the model to produce outputs that share low-level pixel similarities with the actual input target face images, i.e.:  
\begin{equation}
    \mathcal{L}^R_{rec} = \|R(x_{d}) - x\|_{L1},
\end{equation}

where $\|\cdot\|_{L1}$ is the L1 norm.

\item The \textbf{perceptual loss} $\mathcal{L}_{lpips}$ is the Learned Perceptual Image Patch Similarity (LPIPS), described in \cite{LPIPS}, and ensures that higher-level perceptual characteristics of the recovered image $R(x_{d})$ are as close as possible to the original target face image $x$. The loss penalizes dissimilarities in feature maps computed from a VGG16 model pretrained on ImageNet \cite{imagenet}. We do not use the linear layers per feature map scale and directly calculate the loss between the feature maps.

\item The \textbf{identity loss} $\mathcal{L}^R_{id}$ is defined in a similar way as in the main manuscript but aims to minimize the difference in the identity information encoded through the identity encoder $I$, i.e.:

\begin{equation}
  \mathcal{L}^R_{id} = 1 - cos(I(x), I(R(x_d))),
  \label{eq:identity_lossR}
\end{equation}
where $cos(\cdot)$ is the cosine similarity.
\end{itemize}

The overall learning objective of $R$ that is minimized during training is then defined as:

\begin{equation}
  \mathcal{L}_{R} = \lambda^R_{id}\mathcal{L}^R_{id} +\lambda^R_{rec}\mathcal{L}^R_{rec} + \mathcal{L}_{lpips},
  \label{eq:overall}
\end{equation}
where $\lambda^R_{id}=5$ and $\lambda^R_{rec}=5$.


\begin{table*}[!t]
\caption{\textbf{Quantitative results for reconstruction attacks} on FaceForensic++~\cite{faceforensics++}. \textbf{Bold} indicates best, and \underline{underline} indicates second best result.\vspace{-2mm}}
\begin{center}
\resizebox{1.0\textwidth}{!}{
\begin{tabular}{cll|ccccccccccccccc:ccc}
\toprule
&\multirow{3}{*}{Model} & \multirow{3}{*}{Type} &&&&&&                                                                                                                                        \multicolumn{8}{c}{Reconstruction Attack Identity Retrieval$\downarrow$}&& \\
&&&                                                                               \multicolumn{3}{c}{CosFace} & \multicolumn{3}{c}{ArcFace} & \multicolumn{3}{c}{FaceNet} & \multicolumn{3}{c}{AdaFace} & \multicolumn{3}{c}{ElasticFace} & \multicolumn{3}{c}{Average} \\
&&             & \tiny $10^{-3}$ & \tiny $10^{-4}$ & \tiny $10^{-5}$ & \tiny $10^{-3}$ & \tiny $10^{-4}$ & \tiny $10^{-5}$ & \tiny $10^{-3}$ & \tiny $10^{-4}$ & \tiny $10^{-5}$ & \tiny $10^{-3}$ & \tiny $10^{-4}$ & \tiny $10^{-5}$ & \tiny $10^{-3}$ & \tiny $10^{-4}$ & \tiny $10^{-5}$ & \tiny $10^{-3}$ & \tiny $10^{-4}$ & \tiny $10^{-5}$ \\
\cmidrule{2-21}
\parbox[t]{1mm}{\multirow{7}{*}{\rotatebox[origin=c]{90}{\normalsize FaceForensic++}}}
&RiDDLE \cite{RIDDLE} \small{(CVPR'23)} &GAN-Inversion                            & \underline{0.51}  & \underline{0.44}  & \underline{0.09}  & \underline{0.49}  & \underline{0.42}  & \underline{0.21}  & \underline{0.37}  & \underline{0.02}    & \textbf{0.00} & \underline{0.50} & \underline{0.42}& \underline{0.24}    & \underline{0.50}    & \underline{0.48}    & \underline{0.30}    & \underline{0.47}    & \underline{0.36}    & \underline{0.17}    \\
&G2Face \cite{G2Face} \small{(TIFS'24)} &Target-Oriented                          & 1.00              & 1.00              & 0.99              & 1.00              & 1.00              & 0.99              & 0.99              & 0.82          & 0.45          & 1.00             & 1.00            & 1.00          & 1.00          & 1.00          & 1.00          & 1.0           & 0.96          & 0.89\\
&FIT \cite{FIT} \small{(ECCV'20)}        &Target-Oriented                         & 1.00              & 1.00              & 1.00              & 1.00              & 1.00              & 1.00              & 1.00              & 0.98          & 0.84          & 1.00             & 1.00            & 1.00          & 1.00          & 1.00          & 1.00          & 1.00          & 1.00          & 0.97\\
&FIVA \cite{FIVA} \small{(ICCV'23)} &Target-Oriented                              & 0.98              & 0.98              & 0.85              & 0.94              & 0.93              & 0.84              & 0.80              & 0.25          & 0.04          & 0.98             & 0.98            & 0.94          & 0.97          & 0.97          & 0.93          & 0.94          & 0.82          & 0.72\\
&FaceDancer \cite{FaceDancer} + ITM \cite{FIVA} \small{(WACV'23)} &Target-Oriented      & 1.00              & 1.00              & 0.99              & 1.00              & 1.00              & 0.99              & 0.97              & 0.74          & 0.40          & 1.00             & 1.00            & 1.00          & 1.00          & 1.00          & 1.00          & 0.99          & 0.95          & 0.87\\
&SimSwap \cite{simswap} + ITM \cite{FIVA} \small{(ICM'21)} &Target-Oriented            & 0.99              & 0.99              & 0.95              & 0.98              & 0.98              & 0.94              & 0.89              & 0.43          & 0.12          & 0.99             & 0.99            & 0.98          & 0.99          & 0.99          & 0.98          & 0.97          & 0.88          & 0.79\\
& \OURSp~(Ours) &Hybrid                                           & \textbf{0.09}     & \textbf{0.07}     & \textbf{0.01}     & \textbf{0.08}     & \textbf{0.05}     & \textbf{0.02}     & \textbf{0.07}     & \textbf{0.00} & \textbf{0.00} & \textbf{0.07}    & \textbf{0.05}   & \textbf{0.02} & \textbf{0.09} & \textbf{0.08} & \textbf{0.04} & \textbf{0.08} & \textbf{0.05} & \textbf{0.02}\\
\bottomrule
\end{tabular}
}\vspace{-2mm}
\end{center}
\label{t:reconstruction_attack_compare}
\end{table*}
\begin{figure*}[!t]
\centering
\includegraphics[width=1.0\textwidth]{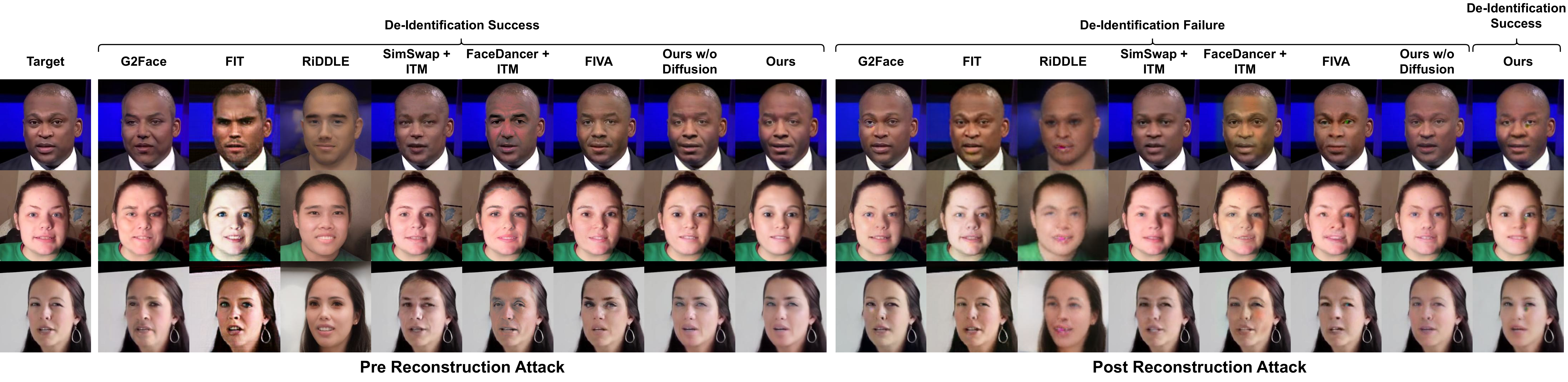}
\caption{\textbf{Qualitative comparisons under reconstruction attacks}. To the left are outputs from  different methods before applying the reconstruction attack. For these samples, the identity is protected from facial recognition. To the right are samples obtained after applying the reconstruction attack. In this case, \OURSp~is the only method that can provide reliable de-identification.}
\label{fig:reconstruction_attacks}
\end{figure*}

\section{Experiments and Results}
\label{s:results}


\noindent\textbf{Experimental Setup:}
When evaluating \OURSp, we consider four key aspects: (1) visual quality/fidelity, (2) attribute retention, (3) identity retrieval, and (4) reconstruction attacks.~For the visual quality of the face region, we use the Fréchet Inception Distance (FID) \cite{fid}.
For attribute retention, we evaluate pose, expression, and gaze preservation capabilities, as well as the mean squared error (MSE) between input and output images, as an attribute agnostic measure for utility preservation. 
In case of pose, we use 3DMM pose coefficients \cite{3DMMEncoder} and a pose estimation network \cite{HOPENET}.
For expression, we rely on 3DMM expression coefficients \cite{3DMMEncoder}. 
When it comes to gaze, we employ a landmark estimation model to regress heatmaps of both, the left and right eyes \cite{AdaptiveWing}. 
We report the mean L2 distance for all attributes.
Identity retrieval is evaluated using five different facial recognition models trained with different SOTA learning objectives, i.e., CosFace \cite{cosface}, ArcFace \cite{arcface}, FaceNet \cite{FACENET}, AdaFace \cite{adaface}, and ElasticFace \cite{elasticface}.
We report the rate of identity retrieval for three False Acceptance Rates (FAR): $10^{-3}$, $10^{-4}$, and $10^{-5}$. 
Reconstruction attack evaluation is conducted by evaluating identity retrieval after the reconstruction attack.
The reconstruction attack models are based on  U-Net, similarly to the generator $G$. 
and are trained to recover target images from the corresponding de-identification results.

\vspace{1.5mm}\noindent\textbf{Datasets:}~We evaluate \OURSp~on three diverse datasets, commonly used when evaluating deidentification models, i.e.: FaceForensic++ \cite{faceforensics++}, CelebA \cite{CelebaA} and Labeled Face in the Wild (LFW) \cite{LFW}.~For FaceForensic++ \cite{faceforensics++}, we follow previous works and sample $10$ frames for each video \cite{FIVA, faceshifter, simswap, hififace}. 
We automatically extract all faces using 5-point landmark detection \cite{retinaface} and compute ArcFace embeddings~\cite{arcface}. 
For videos with multiple people, we split them into separate identities.~Identity retrieval for FaceForensic++ and CelebA is done by searching for a match across the entire dataset, whereas for LFW \cite{LFW} the genuine pair protocol is used.

\vspace{1.5mm}\noindent\textbf{Implementation Details}
The complete \OURS~model and all ablated variants are trained with the same configuration. We use AdamW \cite{AdamW} with a learning rate of $5e^{-5}$, $\beta_0 = 0.5$, $\beta_1 = 0.99$, weight decay of $2e^{-2}$, and epsilon value $\epsilon = 1e^{-8}$. The batch size is 10. Gradients are clipped at a max norm of 1 for all networks. The models are trained in mixed precision, autocasting to float16. We use a set seed (42 in our case) for all relevant packages used (e.g. PyTorch, Numpy). The models are learned in a $256\times256$ resolution and for $50,000$ iterations. The VGGFace2 \cite{vggface2} dataset is used for training, and we randomly load images for each batch.

When training the diffusion model $\epsilon_\theta$, we use a similar setting as with the training of \OURS. Specifically, we adopt AdamW \cite{AdamW} with a learning rate of $1e^{-4}$, $\beta^{opt}_0 = 0.95$, $\beta^{opt}_1 = 0.999$, weight decay of $1e^{-6}$, and epsilon value $\epsilon = 1e^{-8}$. The batch size is 10. Gradients are clipped at a max norm of 1 for all networks. The diffusion model $\epsilon_\theta$ is trained in mixed precision, autocasting to float16. We use a set seed (42 in our case) for all relevant packages used (e.g., PyTorch, Numpy). The diffusion model $\epsilon_\theta$ is trained in a $256\times256$ resolution. The diffusion model $\epsilon_\theta$ is trained for 150000 iterations. The VGGFace2 \cite{vggface2} dataset is used for training, and we randomly load images for each batch. The DDPM schedule \cite{DDPM} for applying noise to the images during training are used with the standard $\tau=1000$ time steps. $\beta_t$ starts at $0.00085$ ($\beta_0$) and ends at $0.012$ ($\beta_{\tau}$), with a linear schedule and leading time step spacing \cite{DDPMLeadingTimesteps}. The diffusion model $\epsilon_\theta$ is of noise $\epsilon$ prediction type.

\begin{figure*}[!t]
\centering
\includegraphics[width=0.9\textwidth]{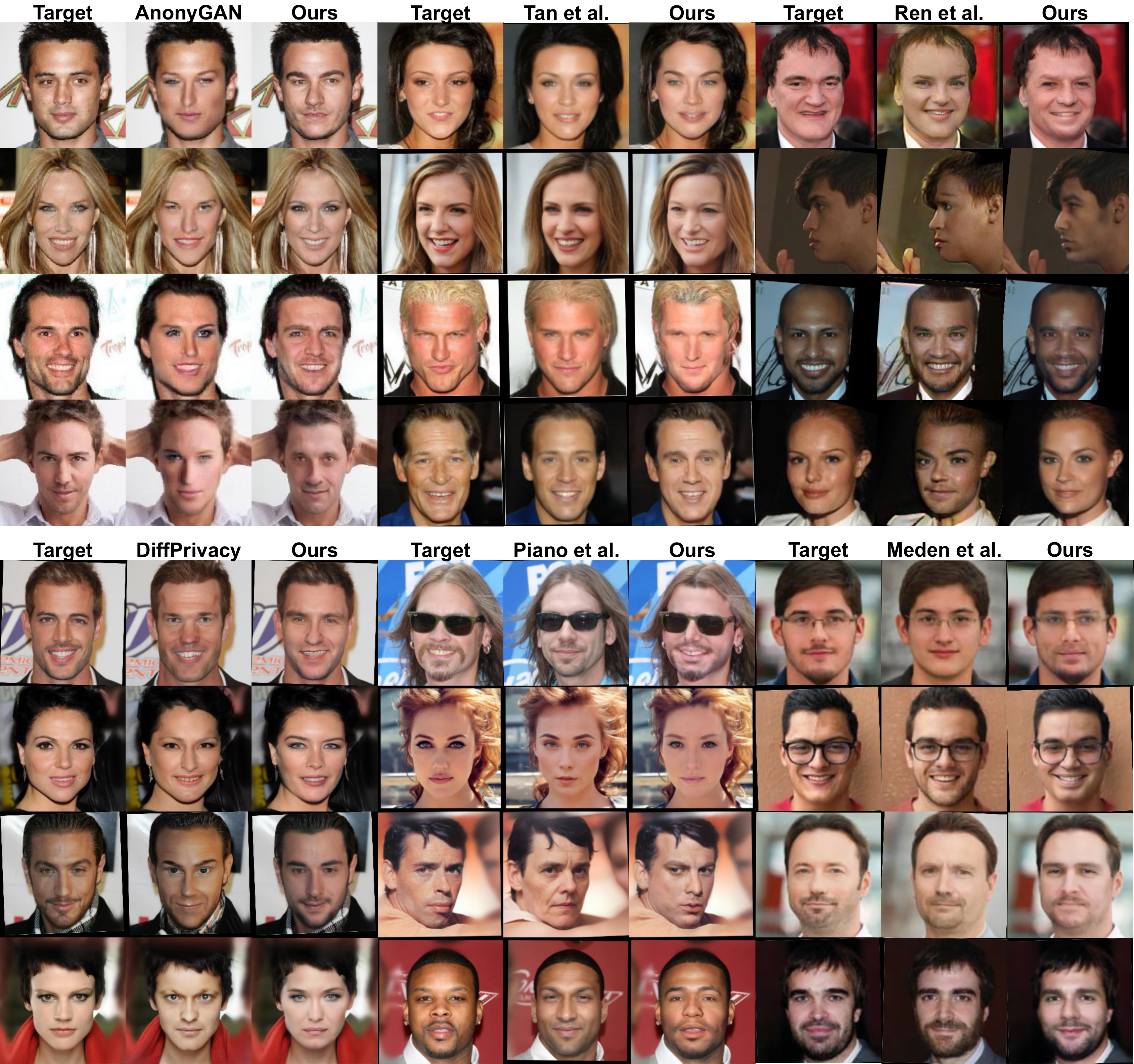}
\caption{\textbf{Qualitative comparisons between \OURSp~and competing de-identification methods}, i.e., with AnonyGAN~\cite{AnonyGAN}, Tan \textit{et al.}~\cite{VisualCoherAnon}, Ren \textit{et al.}~\cite{LEARNINGTOANON}, DiffPrivacy~\cite{DiffusionAnon3}, Piano \textit{et al.}~\cite{DiffusionAnon1} and Meden \textit{et al.}~\cite{DE-ID-VITOMIR}. Note how HYDRO is able to convincingly de-idenitfy the target faces even in challenging scenarios, e.g., in profile view or with occlusions due to glasses.}
\label{fig:qual_compare_external}
\end{figure*}

\subsection{Quantitative Results}
\label{s:quant}
In Tab.~\ref{t:compare}, we report quantitative results for \OURS~ in comparison to other SOTA face de-identification methods on all three datasets.~\OURSp~considerably outperforms all previous works by substantially reducing the FID score, thereby attaining the highest degree of image fidelity.
Together with RiDDLE \cite{RIDDLE} and FIVA \cite{FIVA}, \OURSp~shows the strongest identity protection by returning the lowest identity retrieval FAR scores on average. 
Thus, \OURSp~achieves close to ideal de-identification performance for all facial recognition models.  
With respect to pose, expression, and gaze retention, \OURSp~attains significant improvement compared to all previous works across all datasets.
In terms of computational load (in GFLOPS), \OURSp~is heavier than the lightest methods, i.e., CIAGAN, FIT, DeepPrivacy and FIVA, mostly due to the included diffusion process.~However, this is a reasonable trade-off considering HYDRO's superior attribute-retention capabilities, fidelity and reconstruction attack-robustness. HYDRO is also the top performer w.r.t. the attribute agnostic utility preservation performance indicator, MSE, where it (in general) ensures comparable MSE scores to FIVA and G2Face, while significantly outperforming the rest.



When evaluating reconstruction-attack resilience, we compare \OURSp~with one GAN-inversion method, RiDDLE \cite{RIDDLE}, and five SOTA target-oriented face de-identification methods, 
G2Face \cite{G2Face}, FIT \cite{FIT}, FIVA \cite{FIVA}, FaceDancer \cite{FaceDancer}, and SimSwap \cite{simswap} (the one used in \cite{MYFACEMYCHOICE} for de-identification).~Note that although FaceDancer \cite{FaceDancer} and SimSwap \cite{simswap} are face swapping methods, we couple both with the Identity Tracking Module (ITM) from \cite{FIVA} to employ them as target-oriented de-identification models.~Tab. \ref{t:reconstruction_attack_compare} shows the calculated results for the reconstruction-attack experiments on FaceForensic++~\cite{faceforensics++}.~The results clearly show that \textbf{successful reconstruction-attacks can easily be executed against all competing models}.~With all considered target-oriented methods, the rate of successful identity retrieval is above $72\%$, while in many cases it is between $99\%$ and $100\%$ at various FAR values.~Even the GAN-Inversion method RiDDLE \cite{RIDDLE} is susceptible to this type of attack.~In contrast, the proposed \OURSp~model exhibits substantial resilience against reconstruction attacks.
In comparison with the second-best model (RiDDLE \cite{RIDDLE}), \OURSp~\textbf{reduces the success rate of reconstruction attacks by} $\mathbf{85.7\%}$ on average, and in comparison with the closest target-oriented model (i.e., FIVA \cite{FIVA}) by $94.1\%$ on average across three FAR scores.
This substantial enhancement in resilience paves the way towards non-reversible face de-identification.~Note that the reconstruction attack model for \OURSp~was trained on samples generated with the diffusion step included, indicating that HYDRO is also robust against cat-and-mouse scenarios.

\begin{figure}[!t]
\centering
\includegraphics[width=1.0\columnwidth]{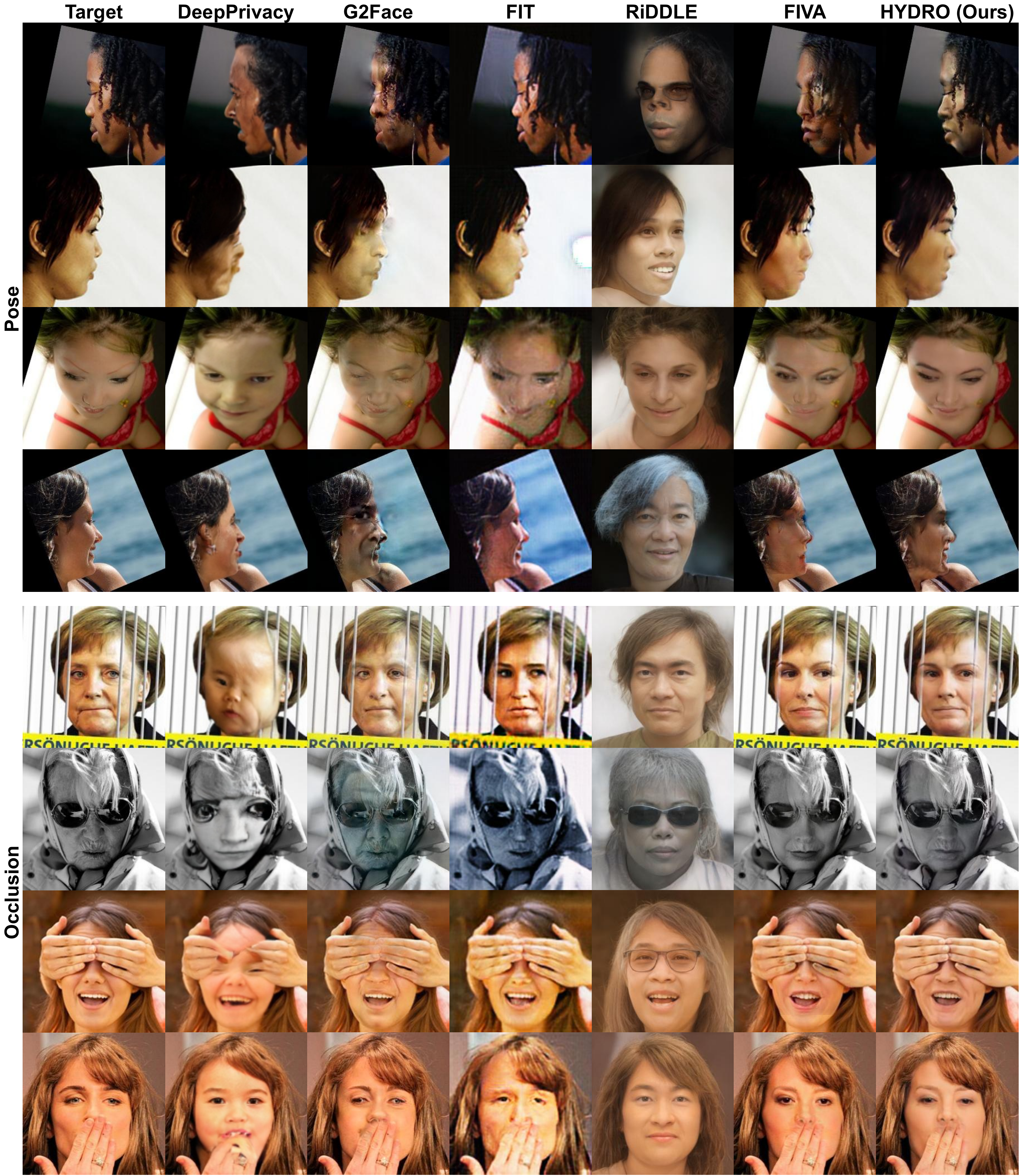}
\caption{\textbf{Qualitative result on images with challenging poses and with occlusions.} The presented results illustrate the capabilities and limitations of HYDRO and competing models with challenging faces. We report comparative results with DeepPrivacy \cite{DEEPPRIVACY}, G2Face \cite{G2Face}, FIT \cite{FIT}, RiDDLE \cite{RIDDLE} and FIVA \cite{FIVA}. Samples are taken from the LS3D-W dataset \cite{ls3dw}.\vspace{-2mm}}
\label{fig:edge_cases_pose_occlusion}
\end{figure}
\begin{figure}[!t]
\centering
\includegraphics[width=1.0\columnwidth]{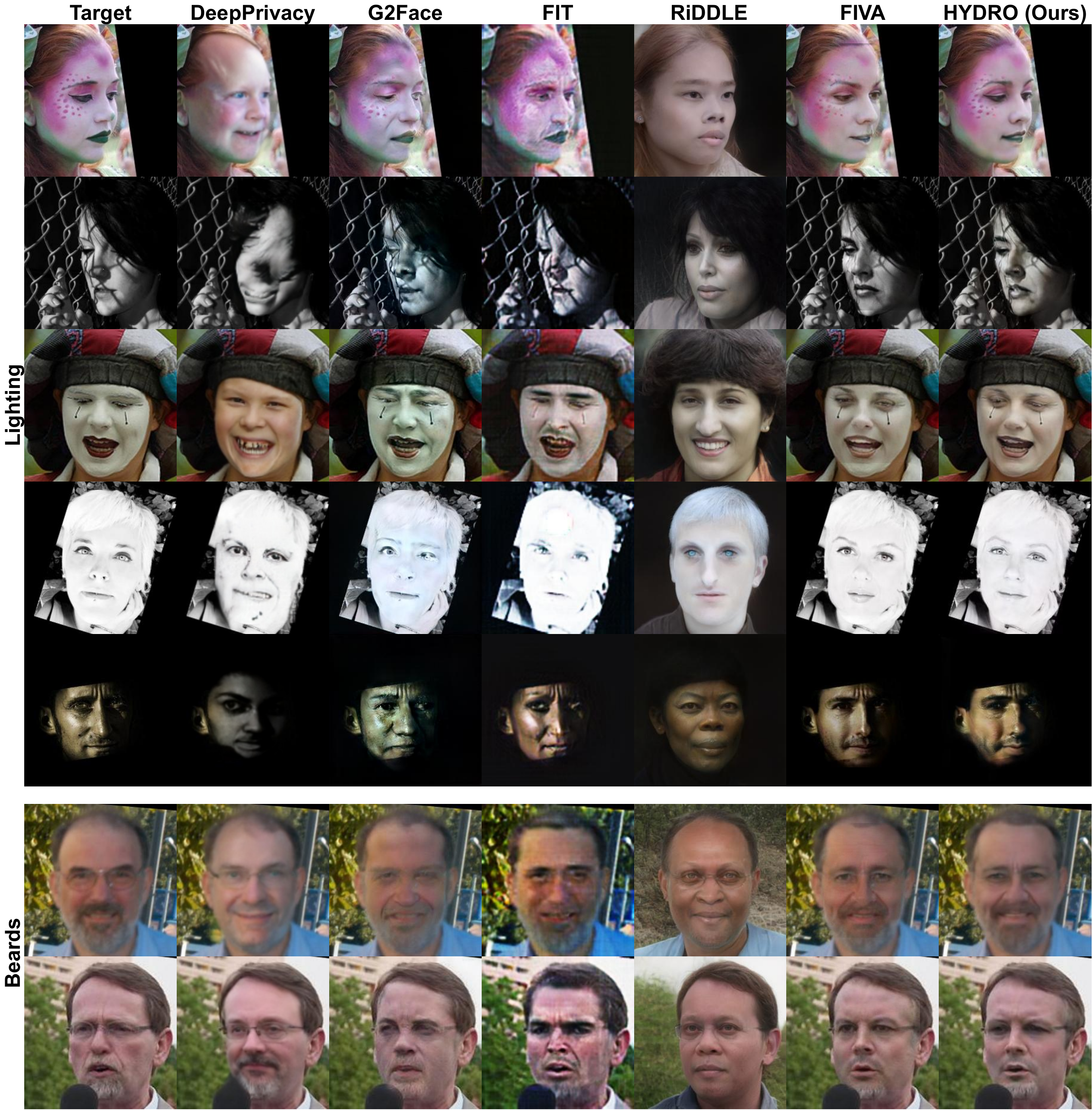}
\caption{\textbf{Qualitative result on images with challenging illumination and faces with beards.}  Comparative results are reported for DeepPrivacy \cite{DEEPPRIVACY}, G2Face \cite{G2Face}, FIT \cite{FIT}, RiDDLE \cite{RIDDLE} and FIVA \cite{FIVA}. Images are taken from  LS3D-W \cite{ls3dw}.}
\label{fig:edge_cases_lighting_beards}
\end{figure}

\subsection{Qualitative Results}
\label{s:qual}

\noindent\textbf{Reconstruction Attacks.} In Fig.~\ref{fig:reconstruction_attacks},~we show qualitative comparisons of the de-identification results and outputs produced thgough the  reconstruction attacks.~As can be seen, G2Face \cite{G2Face}, FIT \cite{FIT}, RiDDLE \cite{RIDDLE}, SimSwap \cite{simswap}+ ITM \cite{FIVA}, and FaceDancer \cite{FaceDancer}+ ITM \cite{FIVA}  encounter difficulties in accurately retaining gender information after de-identification, while FIT \cite{FIT}  additionally alters the overall illumination of the images.
When looking at the reconstruction attacks, we observe that the recovered images based on G2Face \cite{G2Face}, FIT \cite{FIT}, SimSwap \cite{simswap} + ITM \cite{FIVA}, and FaceDancer \cite{FaceDancer} + ITM \cite{FIVA}  are visually very similar to the original targets. 
In the case of FIVA \cite{FIVA}, RiDDLE \cite{RIDDLE}, and \OURSp~without diffusion, the reconstruction is less similar to the initial target, yet the images still contain information sufficient to facilitate a precise match within the dataset.~When it comes to \OURSp, the reconstruction attack is not able to reliably retrieve the original identity.

\noindent\textbf{Additional SOTA Comparisons.} In Fig.~\ref{fig:qual_compare_external}, we further compare \OURSp~with additional strong baselines from the literature. As can be seen, AnonyGAN \cite{AnonyGAN} struggles with symmetry and often times generates less convincing results.
DiffPrivacy \cite{DiffusionAnon3} and Piano \textit{et al.}~\cite{DiffusionAnon1} are both diffusion-based, with computationally expensive pipelines in terms of model sizes and sampling steps. Their visual quality is comparable to \OURSp. However, Piano \textit{et al.}~\cite{DiffusionAnon1} still alter the expression to some extent, while also changing the background. 
Tan \textit{et al.}~\cite{VisualCoherAnon} produce blurry images that affect the background and exhibit problems with gaze retention. 
Ren \textit{et al.}~\cite{LEARNINGTOANON} produce outputs with a significant change in color and visual semantics while introducing artifacts around the borders, and Meden \textit{et al.}~\cite{DE-ID-VITOMIR} lose part of the correspondence with the target image due to the GAN-inversion process.

\begin{figure}[!t]
\centering
\includegraphics[width=1.0\columnwidth]{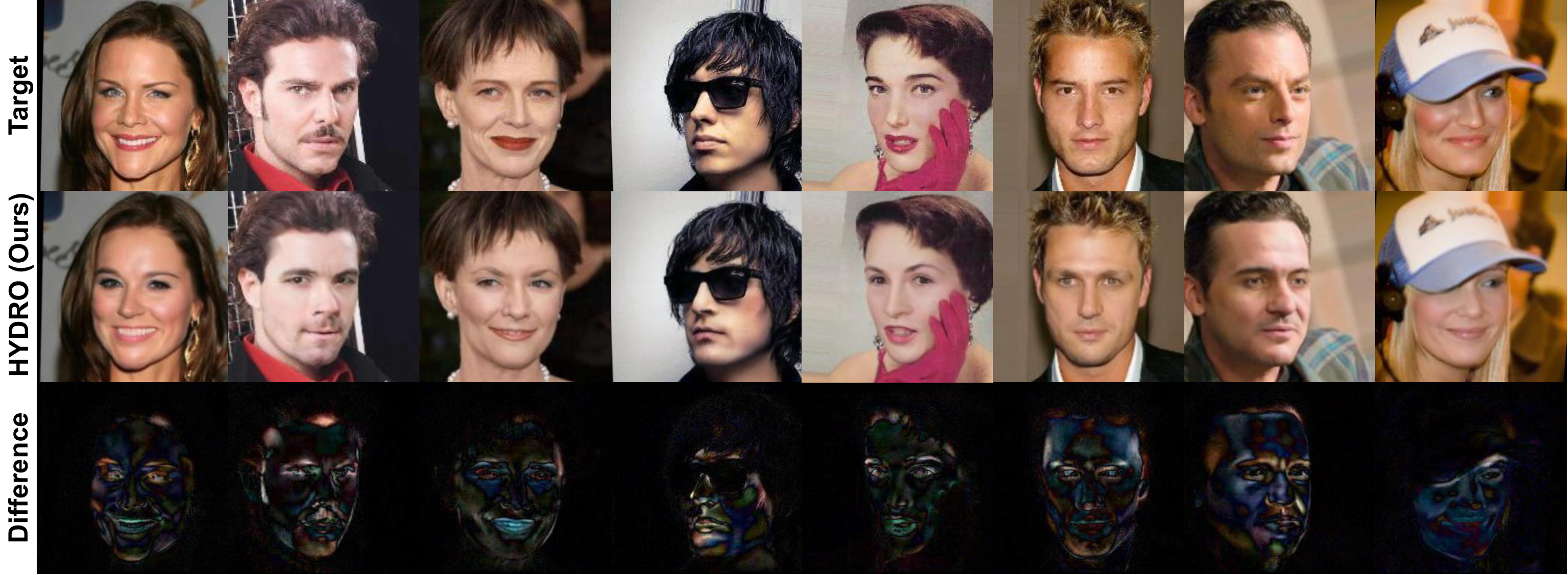}
\caption{\textbf{Illustration of attribute agnostic utility preservation.} The figure shows the absolute difference between the input and output images produced by HYDRO, highlighting the localized changes of the face. Note that HYDRO only changes ares within the facial regions. Attributes, such as hairstyle, image background, hair color, eyeglasses, gloves, hats or clothing, are hence preserved by default.\vspace{-2mm}}
\label{fig:difference}
\end{figure}

\vspace{1.5mm}\noindent\textbf{Edge Cases.} To futher explore the capabilities of HYDRO and competing models with challenging image characteristics and investigate model limitations, we show in Fig. \ref{fig:edge_cases_pose_occlusion} and Fig. \ref{fig:edge_cases_lighting_beards}, example results with face in extreme and less common pose configurations, with occlusions, captured in variable lighting conditions or with the presence of beards. In these comparisons, we consider $5$ state-of-the-art competitors, i.e., DeepPrivacy \cite{DEEPPRIVACY}, G2Face \cite{G2Face}, FIT \cite{FIT}, RIDDLE \cite{RIDDLE} and FIVA \cite{FIVA}. The following observations can be made with respect to the studied image characteristics:
\begin{itemize}[noitemsep,leftmargin=*]
    \item \textbf{Poses:} We observe from the upper part of Fig.~\ref{fig:edge_cases_pose_occlusion} that challenging pose configurations represent an obvious limitation that is prevalent with HYDRO and with the considered SOTA models.
    As can be seen, HYDRO deals with these edge cases better than previous works. DeepPrivacy is inconsistent, G2Face \cite{G2Face} produces significant artifacts, FIT often preserves the original identity instead of performing de-identification, and RiDDLE is limited by the capabilities of the StyleGAN generator. FIVA~\cite{FIVA} achieves comparable results to HYDRO, while HYDRO deals with the edge cases slightly better. However, HYDRO still produces noticeable artifacts and slight inconsistencies in some cases.
    \item \textbf{Occlusions:} A similar limitation is when the occlusion covers a significant portion of the face. Often, HYDRO can deal with this to a greater extent than the competing models, but generally, we still see minute artifacts, such as fading objects, when the hand covers a large portion of the face, as depicted in Fig. \ref{fig:edge_cases_pose_occlusion} -  Occlusion. Nonetheless, even under considerable occlusions, HYDRO produces highly competitive high-fidelity results.
    \item \textbf{Lighting and Beards:} In the upper part of Fig.~\ref{fig:edge_cases_lighting_beards}, we show example de-identification results for challenging lighting conditions and faces with makeup. While FIVA, for example, produces comparable quality visual de-identification results with the presented examples as HYDRO, the rest of the considered models performs worse and struggles to generate realistic results with attributes that resemble the input face image. HYDRO and FIVA, on the other hand, lead to high-fidelity results that are true to the input images and preserve shadows, makeup and other important image attributes. Similar observations can also be made for face with beards. Here, FIVA and HYDRO are again the most convincing models, producing the least amounts of artifacts, while offering a good trade-off between privacy protection, attribute preservation and image fidelity.   
\end{itemize}

\vspace{1.5mm}\noindent\textbf{Utility Preservation Beyond Single Attributes.} In Fig. \ref{fig:difference}, we present a qualitative analysis of the attribute retention capabilities of HYDRO in an attribute agnostic manner. Specifically, we show the absolute difference between the target input images and output images generated by HYDRO. The presented examples demonstrate that HYDRO makes changes that are localized on the facial area only. Therefore, image/facial attributes, such as hairstyle, background appearance or accessories (sunglasses, hats, gloves, clothings, etc.) are preserved by design. Similarly to Fig.~\ref{fig:edge_cases_pose_occlusion}, the presented examples also show how HYDRO is able to produce realistic deidentification results even in the presence of occluding objects.

\begin{figure}[!t]
\centering
\includegraphics[width=0.5\textwidth]{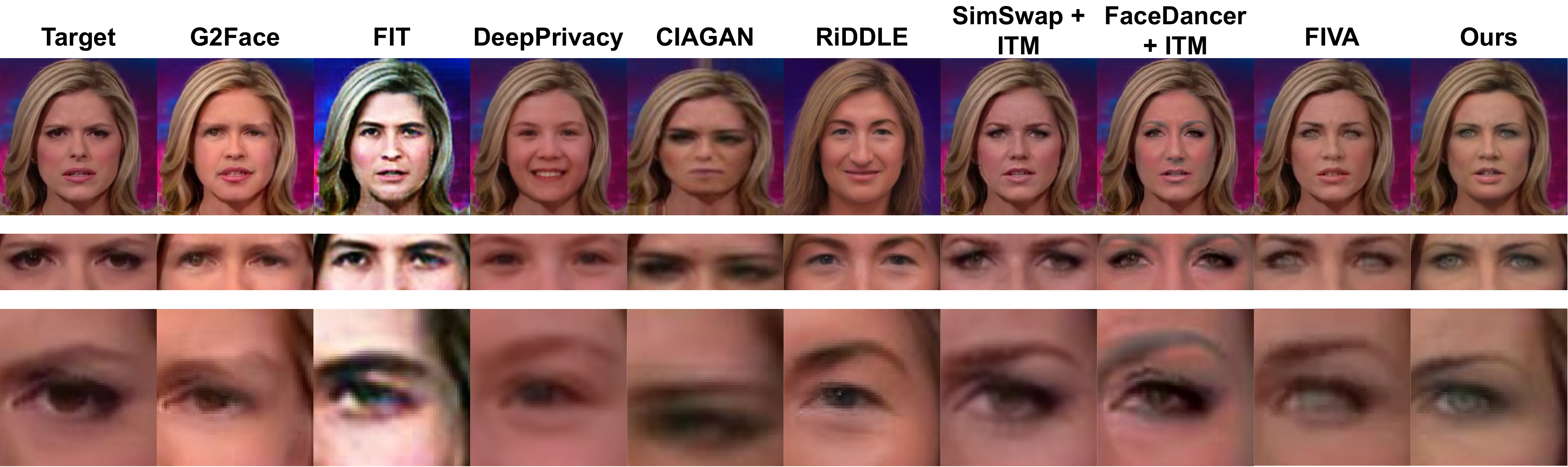}
\caption{\textbf{Qualitative comparisons for the eyes.} The input (target) image is the sample without any de-identification.}
\label{fig:compare_eyes}
\end{figure}
\begin{table}[t!]
\caption{\textbf{Quantitative ablations} on FaceForensic++~\cite{faceforensics++}, LFW~\cite{LFW}, and CelebA~\cite{CelebaA}. \textbf{Bold} indicates best, \underline{underline} indicates second best.~We also report the number of GFLOPS and inference speed in [ms] as calculated on an Nvidia A6000.}
\begin{center}
\resizebox{0.99\columnwidth}{!}{
\begin{tabular}{cl|ccccc:cc}
\toprule
&& FID$\downarrow$ & Pose$\downarrow$ & Pose$\downarrow$ & Exp$\downarrow$ & Gaze$\downarrow$ & GFLOPS$\downarrow$ & Inference$\downarrow$ \\

&Model             & & \cite{3DMMEncoder} & \cite{HOPENET} & \cite{3DMMEncoder} & \cite{AdaptiveWing} & & [ms]\\
\cmidrule{2-9}
\parbox[t]{2mm}{\multirow{6}{*}{\rotatebox[origin=c]{90}{\normalsize FaceForensic++}}}
&Baseline                                         & 5.59            & \textbf{0.03} & 1.86            & \textbf{2.17}   & 6.81            & \textbf{323.09}& \textbf{36.45}\\
&+Attention                                       & 8.81            & \textbf{0.03} & \textbf{1.77}   & 2.29            & 6.95            & \underline{339.24}& \underline{43.45}\\
&+Gaze loss                                       & \underline{4.50}& \textbf{0.03} & 1.79            & 2.24            & 6.51            & \underline{339.24}& \underline{43.45}\\
&+Eye Discrimination                              & 6.03            & \textbf{0.03} & 1.89            & 2.32            & 6.60            & \underline{339.24}& \underline{43.45}\\
&+Eye Similarity loss                             & 7.32            & \textbf{0.03} & 1.80            & 2.30            & \underline{6.24}& \underline{339.24}& \underline{43.45}\\
&+Diffusion (\OURS)                               & \textbf{3.55}   & \textbf{0.03} & \textbf{1.77}   & \underline{2.22}& \textbf{6.00}   & 654.19& 62.05\\
\cmidrule{2-9}
\parbox[t]{2mm}{\multirow{6}{*}{\rotatebox[origin=c]{90}{\normalsize CelebA}}}
&Baseline                                         & 1.88            & \textbf{0.03} & 2.31            & \textbf{2.13}   & 6.96            &\textbf{323.09}& \textbf{36.45}\\
&+Attention                                       & 1.90            & \textbf{0.03} & 2.29            & 2.24            & 7.10            & \underline{339.24}& \underline{43.45}\\
&+Gaze loss                                       & \textbf{0.84}   & \textbf{0.03} & 2.30            & 2.18            & 6.54            & \underline{339.24}& \underline{43.45}\\
&+Eye Discrimination                              & 2.21            & \textbf{0.03} & 2.27            & 2.24            & 6.60            & \underline{339.24}& \underline{43.45}\\
&+Eye Similarity loss                             & 1.85            & \textbf{0.03} & \underline{2.25}& 2.22            & \underline{6.36}& \underline{339.24}& \underline{43.45}\\
&+Diffusion (\OURS)                               & \underline{1.14}& \textbf{0.03} & \textbf{2.14}   & \underline{2.16}& \textbf{6.13}   & 654.19& 62.05\\
\cmidrule{2-9}
\parbox[t]{2mm}{\multirow{6}{*}{\rotatebox[origin=c]{90}{\normalsize LFW}}}
&Baseline                                         & 3.50            & \textbf{0.03} & 2.47            & \textbf{2.26}   & 7.25            & \textbf{323.09}& \textbf{36.45}\\
&+Attention                                       & 3.56            & \textbf{0.03} & 2.39            & 2.36            & 7.42            & \underline{339.24}& \underline{43.45}\\
&+Gaze loss                                       & 4.07            & \textbf{0.03} & 2.41            & 2.31            & 6.83            & \underline{339.24}& \underline{43.45}\\
&+Eye Discrimination                              & \underline{3.33}& \textbf{0.03} & 2.41            & 2.36            & 6.84            & \underline{339.24}& \underline{43.45}\\
&+Eye Similarity loss                             & 3.36            & \textbf{0.03} & \underline{2.35}& 2.37            & \underline{6.59}& \underline{339.24}& \underline{43.45}\\
&+Diffusion (\OURS)                               & \textbf{3.08}   & \textbf{0.03} & \textbf{2.31}   & \underline{2.28}& \textbf{6.34}   & 654.19& 62.05\\
\bottomrule
\end{tabular}
}
\end{center}
\label{t:ablations}
\end{table}

\begin{table*}[!t!]
\caption{\textbf{Quantitative ablations for reconstruction attacks for \OURSp~ on FaceForensic++~\cite{faceforensics++}.} \textbf{Bold} indicates best, and \underline{underline} indicates second best. $t$ is the time step passed to the diffusion model, $\sigma_t$ is the noise ratio applied -- see Eq.~\eqref{eq:dif_noise}.}
\begin{center}
\resizebox{1.0\textwidth}{!}{
\begin{tabular}{ccc|ccccc:ccccccccccccccc:ccc}
\toprule
&&&&&&&&&&&&&                                                                                                                                        \multicolumn{8}{c}{Reconstruction Attack Identity Retrieval$\downarrow$}&& \\
&&& FID$\downarrow$ & Pose$\downarrow$ & Pose$\downarrow$ & Exp$\downarrow$ & Gaze$\downarrow$ & \multicolumn{3}{c}{CosFace} & \multicolumn{3}{c}{ArcFace} & \multicolumn{3}{c}{FaceNet} & \multicolumn{3}{c}{AdaFace} & \multicolumn{3}{c}{ElasticFace} & \multicolumn{3}{c}{Average} \\

& $\sigma_t$ & $t$            & & \cite{3DMMEncoder} & \cite{HOPENET} & \cite{3DMMEncoder} & \cite{AdaptiveWing} & \tiny $10^{-3}$ & \tiny $10^{-4}$ & \tiny $10^{-5}$ & \tiny $10^{-3}$ & \tiny $10^{-4}$ & \tiny $10^{-5}$ & \tiny $10^{-3}$ & \tiny $10^{-4}$ & \tiny $10^{-5}$ & \tiny $10^{-3}$ & \tiny $10^{-4}$ & \tiny $10^{-5}$ & \tiny $10^{-3}$ & \tiny $10^{-4}$ & \tiny $10^{-5}$ & \tiny $10^{-3}$ & \tiny $10^{-4}$ & \tiny $10^{-5}$ \\
\cmidrule{2-26}
\parbox[t]{2mm}{\multirow{8}{*}{\rotatebox[origin=c]{90}{\normalsize FaceForensic++}}}
&None   & None  & 7.32              & 0.03          & 1.80             & 2.30         & 6.24            & 0.99            & 0.99            &  0.91            & 0.96            & 0.95            & 0.87          & 0.91            & 0.43          & 0.10          & 0.98           & 0.98          & 0.96          & 0.99          & 0.99          & 0.99          & 0.97          & 0.87          & 0.77           \\
&0.0414 & None  & 144.13            & 0.05          & 5.58             & 2.32         & 6.42            & 0.03            & 0.02            &  0.00            & 0.02            & 0.01            & 0.00          & 0.08            & 0.01          & 0.00          & 0.04           & 0.02          & 0.01          & 0.03          & 0.03          & 0.01          & 0.04          & 0.02          & 0.00          \\
\cdashline{2-26}
&0.0414 & 1     & 3.91              & \textbf{0.03} & \textbf{1.73}    & \textbf{2.21}& \textbf{5.97}   & 0.17            & 0.13            &  \underline{0.02}& 0.14            & 0.10            & 0.04          & 0.11            & 0.01          & 0.00            & 0.14            & 0.11            & 0.06            & 0.17            & 0.15            & 0.07          & 0.14 & 0.10 & 0.04 \\
&0.0509 & 2     & 4.65              & \textbf{0.03} & \textbf{1.73}    & \textbf{2.21}& \textbf{5.97}   & 0.15            & 0.12            &  \underline{0.02}& 0.13            & 0.09            & 0.03          & 0.10            & 0.01          & 0.00            & 0.13            & 0.10            & 0.05            & 0.16            & 0.14            & 0.06          & 0.13 & 0.09 & 0.03 \\
&0.0589 & 3     & 4.65              & \textbf{0.03} & 1.74             & \textbf{2.21}& 5.98            & 0.14            & 0.11            &  \underline{0.02}& 0.12            & 0.08            & 0.03          & 0.10            & 0.01          & 0.00            & 0.12            & 0.09            & 0.04            & 0.15            & 0.13            & 0.06          & 0.12 & 0.08 & 0.03 \\
&0.0661 & 4     & 4.49              & \textbf{0.03} & 1.74             & \textbf{2.21}& 5.98            & 0.13            & 0.10            &  \underline{0.02}& 0.11            & 0.08            & 0.03          & 0.09            & 0.01          & 0.00            & 0.11            & 0.09            & 0.04            & 0.14            & 0.12            & 0.06          & 0.12 & 0.08 & 0.03 \\
&0.0727 & 5     & \underline{3.65}  & \textbf{0.03} & 1.75             & \textbf{2.21}& 5.98            & 0.13            & 0.10            &  \underline{0.02}& 0.11            & 0.08            & \textbf{0.02} & 0.09            & 0.01          & 0.00            & 0.11            & 0.08            & 0.04            & 0.13            & 0.11            & 0.05          & 0.11 & 0.07 & 0.03 \\
&0.1001 & 10    & \textbf{3.55}     & \textbf{0.03} & 1.77             & 2.22         & 6.00            & \underline{0.11}& \underline{0.08}&  \textbf{0.01}   & \underline{0.09}& \underline{0.06}& \textbf{0.02} & \underline{0.08}& \textbf{0.00} & \textbf{0.00}   & \underline{0.09}& \underline{0.07}& \underline{0.03}& \underline{0.11}& \underline{0.10}& \textbf{0.04} & \underline{0.10} & \underline{0.06} & \textbf{0.02} \\
&0.1429 & 20    & 4.10              & \textbf{0.03} & 1.79             & 2.22         & 6.03            & \textbf{0.09}   & \textbf{0.07}   &  \textbf{0.01}   & \textbf{0.08}   & \textbf{0.05}   & \textbf{0.02} & \textbf{0.07}   & \textbf{0.00} & \textbf{0.00}   & \textbf{0.07}   & \textbf{0.05}   & \textbf{0.02}   & \textbf{0.09}   & \textbf{0.08}   & \textbf{0.04} & \textbf{0.08} & \textbf{0.05} & \textbf{0.02} \\
\bottomrule
\end{tabular}
}
\end{center}
\label{t:reconstruction_attack_ablations}
\end{table*}

\vspace{1.5mm}\noindent\textbf{Eye-Quality Analysis.} Fig.~\ref{fig:compare_eyes} depicts qualitative eye comparisons, with a particular focus on the enhanced visual quality and symmetry of the eye regions in comparison to previous studies.
G2Face \cite{G2Face}, FIT \cite{FIT}, DeepPrivacy \cite{DEEPPRIVACY}, CIAGAN \cite{CIAGAN}, FaceDancer \cite{FaceDancer}, and FIVA \cite{FIVA} commonly demonstrate blurry eyes or missing pupils. SimSwap \cite{simswap} manages to capture the pupil, but exhibits slightly worse quality and eye symmetry compared to \OURSp.
Although RiDDLE \cite{RIDDLE} generates high fidelity eyes, it fails to preserve eye gaze and visual semantics as shown in other cases (See Fig. \ref{fig:reconstruction_attacks} and Tab. \ref{t:compare}).

\subsection{Ablation Studies}
\label{ss:ablation_study}

\begin{figure}[!t]
\centering
\includegraphics[width=0.5\textwidth]{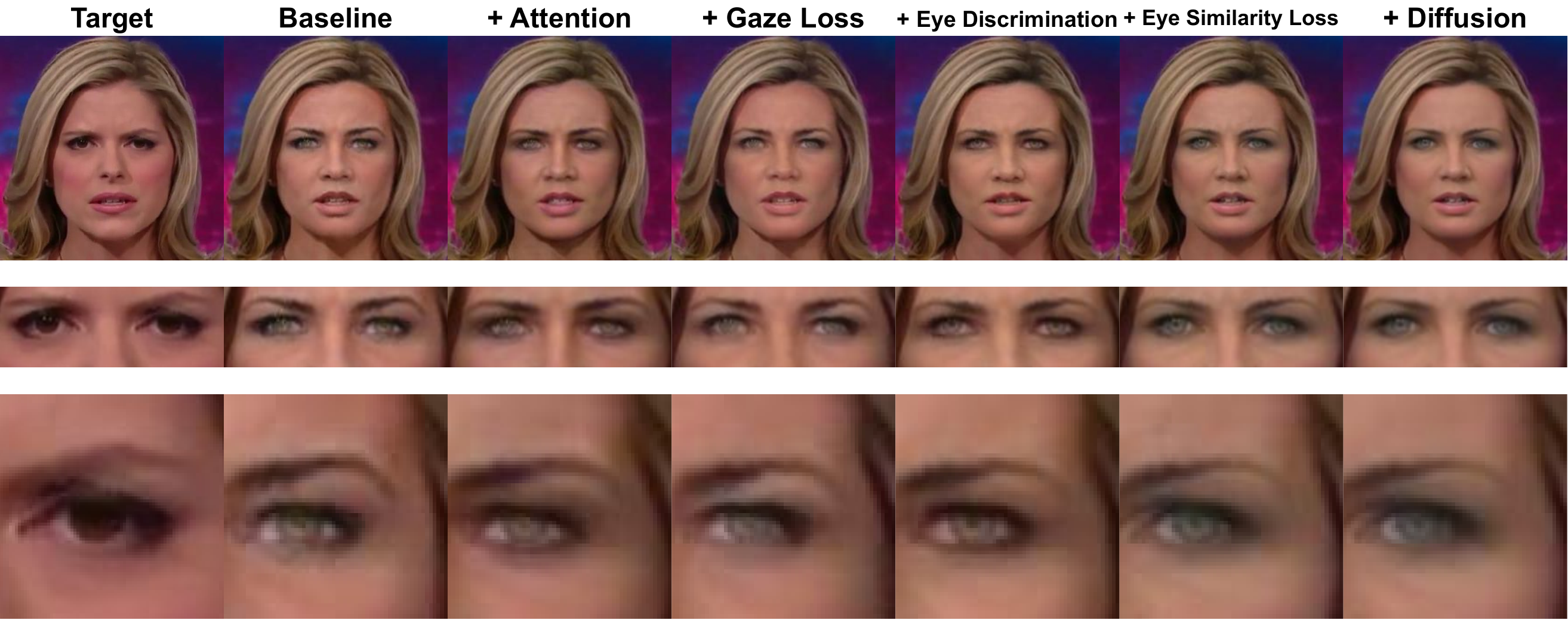}
\caption{\textbf{Qualitative results for the ablation study}, showing the impact of the Eye Similarity Discriminator. 
}
\label{fig:ablations}
\end{figure}

\vspace{1.5mm}\noindent\textbf{Architectural Components.} Tab.~\ref{t:ablations} shows results of the first ablation study, where different \OURSp~components were ablated for the analysis.~We start with a Baseline U-Net generator and then gradually add: attention blocks, the gaze loss, the eye discriminator, the eye-similarity loss, and the one-step backward diffusion process. 
We observe a significant increase in gaze retention, when adding the eye-similarity loss, and general attribute preservation with the inclusion of the diffusion recovery. 
We also see a strong improvement in visual quality without negative effects on de-identification. The improvement in gaze retention and maintained FID show that the proposed \ESD~can significantly improve the quality of the synthetic eyes, as qualitatively shown in Fig.~\ref{fig:ablations}. We utilize and evaluate a gaze loss, as used in \cite{GHOST}, and show further improvements with \ESD. Overall, the reported ablation results point to the importance of all \OURSp~components that reasonably and consistently improve different de-identification trade-offs, when incorporated with the baseline model.

\vspace{1.5mm}\noindent\textbf{Reconstruction Attack.} In Tab.~\ref{t:reconstruction_attack_ablations}, we show comparisons and results for reconstruction attacks against \OURSp~for different noise and diffusion configurations. 
When excluding noise or any diffusion step, the reconstruction model manages to retrieve the original identity $99\%$ of the time for many FAR thresholds and face recognition model combinations. If the noise strength is increased to $0.0414$ (see Eq.~\eqref{eq:dif_noise}), it is possible to prevent reconstruction attacks, but at the cost of degrading data fidelity, as shown by the decreased FID score. Attribute retention also degrades with stronger noise levels.

As shown in Tabs.~\ref{t:reconstruction_attack_ablations} and~\ref{t:ablations}, applying the diffusion model not only mitigates the risk of reconstruction attacks, but also enhances the visual quality and attribute retention capabilities of HYDRO. 
However, the diffusion step introduces additional computational overhead, as evidenced by the high GFLOPS and inference time reported in Tab.~\ref{t:ablations}.

\vspace{1.5mm}\noindent\textbf{Loss Ablation Studies.} HYDRO employs a diverse array of loss functions to ensure the retention of attributes while simultaneously manipulating the underlying identity. In this section, ablations are provided for the attribute retention losses to gauge their contribution, except for the gaze loss $\mathcal{L}_{gze}$, which is discussed and evaluated in great detail in the main manuscript. Tab. \ref{t:sm:loss_ablations} shows the results obtained, when different losses are disabled. 
In general, all losses contribute either to attribute retention or to the fidelity (i.e., the FID score). 
The feature-similarity loss $\mathcal{L}_{fsr}$ \cite{FIVA, FaceDancer} is a crucial component in both attribute retention and maintaining a low FID score. Similarly, the L1 distance loss $\mathcal{L}_{shp}$ between expression and pose coefficients \cite{3DMMEncoder} contributes to attribute retention across all datasets and results in a low FID score for FaceForensic++. Disabling the mask regularization term $\mathcal{L}_{msk}$ or the L1 reconstruction loss $\mathcal{L}_{rec}$ results in improved FID scores but reduced attribute retention for CelebA and LFW. Conversely, the FID score and gaze retention get  worse for FaceForensic++.
In general, there is a trade-off between the FID score and attribute retention, and with each loss this trade-off gap is being reduced, suggesting that all losses importantly contribute to the overall performance of HYDRO.

\begin{table}[t!]
\caption{\textbf{Quantitative loss ablations} on FaceForensic++~\cite{faceforensics++}, LFW~\cite{LFW}, and CelebA~\cite{CelebaA}. \textbf{Bold} indicates best, \underline{underline} indicates second best.~We show results for HYDRO without the diffusion process. For clarity, we also show the results of the full HYDRO pipeline.}
\begin{center}
\resizebox{0.99\columnwidth}{!}{
\begin{tabular}{cl|ccccc}
\toprule
&& FID$\downarrow$ & Pose$\downarrow$ & Pose$\downarrow$ & Exp$\downarrow$ & Gaze$\downarrow$\\

&Model             & & \cite{3DMMEncoder} & \cite{HOPENET} & \cite{3DMMEncoder} & \cite{AdaptiveWing}\\
\cmidrule{2-7}
\parbox[t]{2mm}{\multirow{6}{*}{\rotatebox[origin=c]{90}{\normalsize FaceForensic++}}}
&HYDRO                                                                     & 3.55  & 0.03 & 1.77 & 2.22 & 6.00 \\
\cdashline{2-7}
&HYDRO $w \backslash o$ Diffusion                                          & \textbf{1.85}   & \textbf{0.03} & \textbf{1.80}   & \textbf{2.30}   & \textbf{6.24} \\
&HYDRO $w \backslash o$ Diffusion $w \backslash o $ $\mathcal{L}_{fsr}$    & 23.72           & 0.06          & 4.18            & 2.92            & 12.98\\
&HYDRO $w \backslash o$ Diffusion $w \backslash o $ $\mathcal{L}_{shp}$    & 11.87           & 0.04          & 2.19            & 2.98            & 7.41 \\
&HYDRO $w \backslash o$ Diffusion $w \backslash o $ $\mathcal{L}_{msk}$    & 8.33            & \textbf{0.03} & \underline{1.92}& \underline{2.32}& \underline{6.59} \\
&HYDRO $w \backslash o$ Diffusion $w \backslash o $ $\mathcal{L}_{rec}$    & \underline{6.91}& \textbf{0.03} & 1.97            & 2.33            & 6.65 \\
\cmidrule{2-7}
\parbox[t]{2mm}{\multirow{6}{*}{\rotatebox[origin=c]{90}{\normalsize CelebA}}}
&HYDRO                                                                     & 1.14  & 0.03 & 2.14 & 2.16 & 6.13 \\
\cdashline{2-7}
&HYDRO $w \backslash o$ Diffusion                                          & 1.85            & \textbf{0.03} & \textbf{2.25}   & \textbf{2.22}   & \textbf{6.36} \\
&HYDRO $w \backslash o$ Diffusion $w \backslash o $ $\mathcal{L}_{fsr}$    & 7.98            & 0.07          & 4.79            & 2.75            & 12.95\\
&HYDRO $w \backslash o$ Diffusion $w \backslash o $ $\mathcal{L}_{shp}$    & 1.85            & 0.05          & 2.69            & 2.79            & 7.34 \\
&HYDRO $w \backslash o$ Diffusion $w \backslash o $ $\mathcal{L}_{msk}$    & \underline{1.57}& 0.04          & \underline{2.35}& 2.24            & \underline{6.64} \\
&HYDRO $w \backslash o$ Diffusion $w \backslash o $ $\mathcal{L}_{rec}$    & \textbf{0.59}   & \textbf{0.03} & 2.41            & \underline{2.23}& 6.68 \\
\cmidrule{2-7}
\parbox[t]{2mm}{\multirow{6}{*}{\rotatebox[origin=c]{90}{\normalsize LFW}}}
&HYDRO                                                                     & 3.08  & 0.03 & 2.31 & 2.28 & 6.34 \\
\cdashline{2-7}
&HYDRO $w \backslash o$ Diffusion                                          & 3.36            & \textbf{0.03} & \textbf{2.35}   & 2.37          & \textbf{6.59} \\
&HYDRO $w \backslash o$ Diffusion $w \backslash o $ $\mathcal{L}_{fsr}$    & 10.87           & 0.06          & 4.85            & 2.98          & 13.34\\
&HYDRO $w \backslash o$ Diffusion $w \backslash o $ $\mathcal{L}_{shp}$    & 2.55            & 0.04          & 2.77            & 2.99          & 7.68 \\
&HYDRO $w \backslash o$ Diffusion $w \backslash o $ $\mathcal{L}_{msk}$    & \underline{2.52}& \textbf{0.03} & 2.49            & \textbf{2.36} & \underline{6.92} \\
&HYDRO $w \backslash o$ Diffusion $w \backslash o $ $\mathcal{L}_{rec}$    & \textbf{2.01}   & \textbf{0.03} & \underline{2.45}& \textbf{2.36} & 7.01 \\
\bottomrule
\end{tabular}
}
\end{center}
\label{t:sm:loss_ablations}
\end{table}

\section{Conclusion}
This paper introduced a novel hybrid face de-identification approach, named \OURSp, that was shown to yield state-of-the-art de-identification and attribute-retention capabilities, while ensuring high-fidelity results and exhibiting unprecedented resilience against reconstruction attacks.

\vspace{1mm}\noindent\textbf{Limitations:} There exists a trade-off between attribute retention and de-identification performance in face de-identification methods.~Although \OURSp~shares this limitation, it reduces the trade-off gap.~While the diffusion step of \OURSp~is critical for ensuring fidelity and resilience to reconstruction attacks, it also increases GFLOPS and the computation time, as seen from the results in Tab. \ref{t:ablations}.

\vspace{1mm}\noindent\textbf{Ethics Statement:} Target-oriented methods are common in face-swapping frameworks \cite{FaceDancer, simswap, hififace, faceshifter}. \OURSp~is trained in a counterfactual approach, which enables it to impose a source identity without the image being convincing \cite{FIVA}. 
This enables the utilization of de-identification functionality, while impeding the feasibility of deep faking.

\include{supplementary_material}
\section*{Acknowledgment}
Co-funded by Vinnova Advanced Digitalization, Cyber Security for Industrial Advanced Digitalization. Project grant no. 2023-02996 and the ARIS Research Program P2-0250(B).


\ifCLASSOPTIONcaptionsoff
  \newpage
\fi



%
{\small
\bibliographystyle{Transactions-Bibliography/IEEEtran.bst}
\bibliography{egbib}
}

%

\begin{IEEEbiography}[{\includegraphics[width=1in,height=1.25in,clip,keepaspectratio]{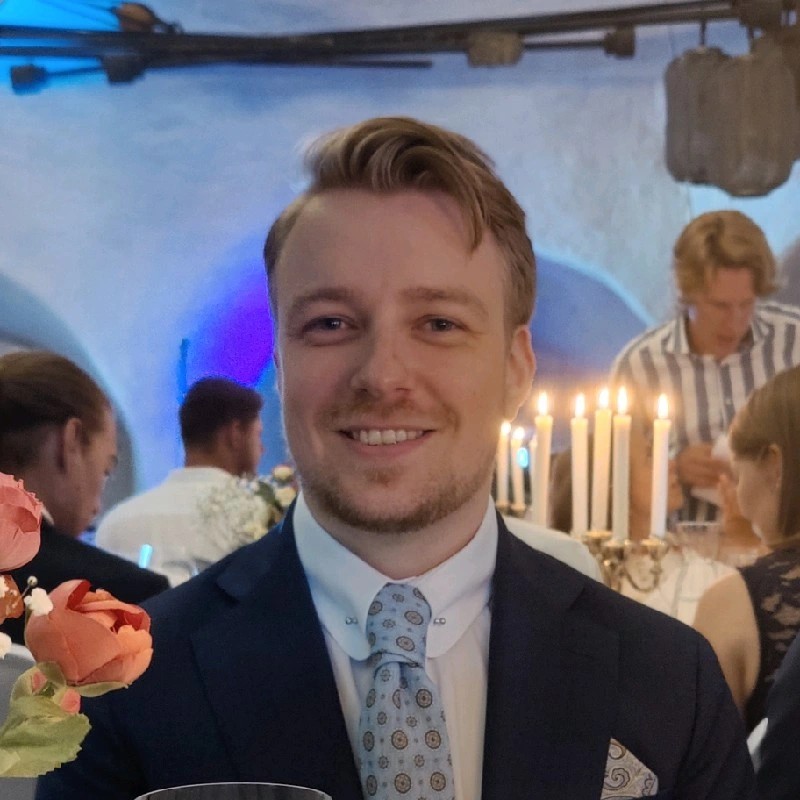}}]{Felix Rosberg}
recieved his M.Sc. in computer science and intelligent systems and Ph.D. in signal processing from Halmstad University, Sweden, in 2020 and 2025, respectively. His research interests include biometrics, generative AI, self-supervised learning, and computer vision for enabling autonomous driving.
\end{IEEEbiography}

\begin{IEEEbiography}[{\includegraphics[width=1in,height=1.25in,clip,keepaspectratio]{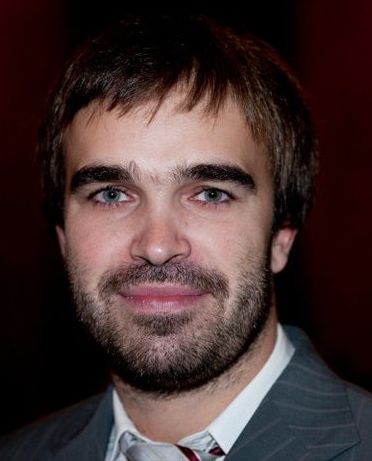}}]{Vitomir Štruc}
is a Full Professor at the University of Ljubljana, Slovenia. His research interests include problems related to biometrics, computer vision, image processing, and machine learning. He (co-)authored more than 200 research papers for leading international peer reviewed journals and conferences in these and related areas. Vitomir is a Deputy Editor-in-Chief for the IEEE Transactions on Information Forensics and Security, a Subject Editor for Elsevier’s Signal Processing and an Associate Editor for Pattern Recognition. He regularly serves on the organizing committees of visible international conferences, including IJCB, FG, WACV and CVPR. He was a Program Co-Chair for IJCB 2023, IEEE FG 2024 and WACV 2025, and currently acts as a Program Chair for IJCB 2025 and General Co-Chair for WACV 2026. Dr. Struc is a Senior member of the IEEE, a member of IAPR, EURASIP, Slovenia’s ambassador for the European Association for Biometrics (EAB) and the former president and current executive committee member of the Slovenian Pattern Recognition Society, the Slovenian member of IAPR. Vitomir is also the current VP Technical Activities for the IEEE Biometrics Council, the secretary of the IAPR Technical Committee on Biometrics (TC4) and a member of the Supervisory Board of the EAB.
\end{IEEEbiography}

\begin{IEEEbiography}[{\includegraphics[width=1in,height=1.25in,clip,keepaspectratio]{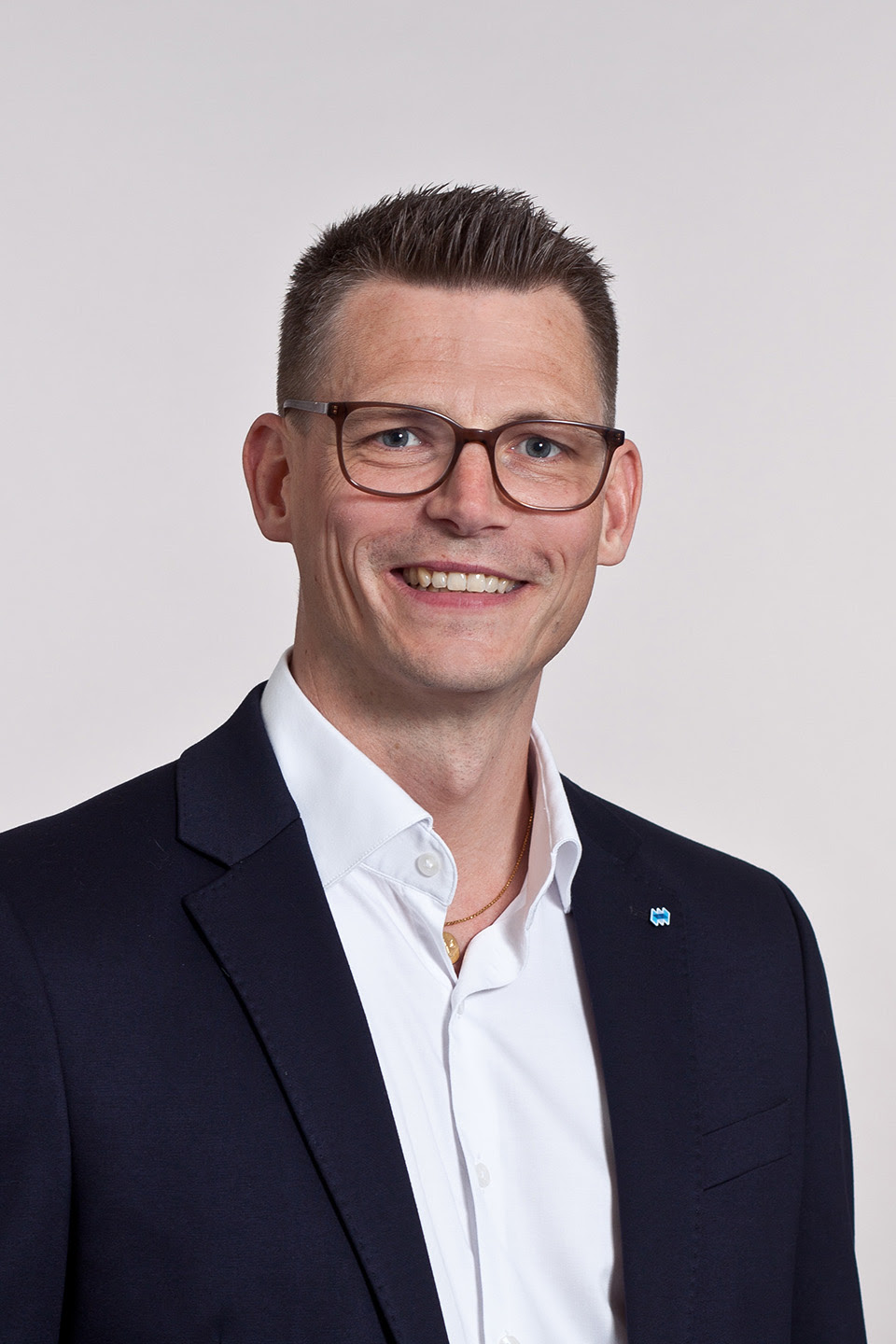}}]{Cristofer Englund}
is a docent and Professor at Halmstad University, Sweden, where he also received his B.Sc. and M.Sc. degrees in electrical engineering and computer science in 2021 and 2023, respectively. He received his PhD in electrical engineering from Chalmers University of Technology, Sweden, in 2007. After several years in the industry and research institute domain, he is currently dean of the School of Information Technology at Halmstad University. His research interests include trustworthy and explainable AI and, computer vision and machine learning for traffic safety applications.
\end{IEEEbiography}

\begin{IEEEbiography}[{\includegraphics[width=1in,height=1.25in,clip,keepaspectratio]{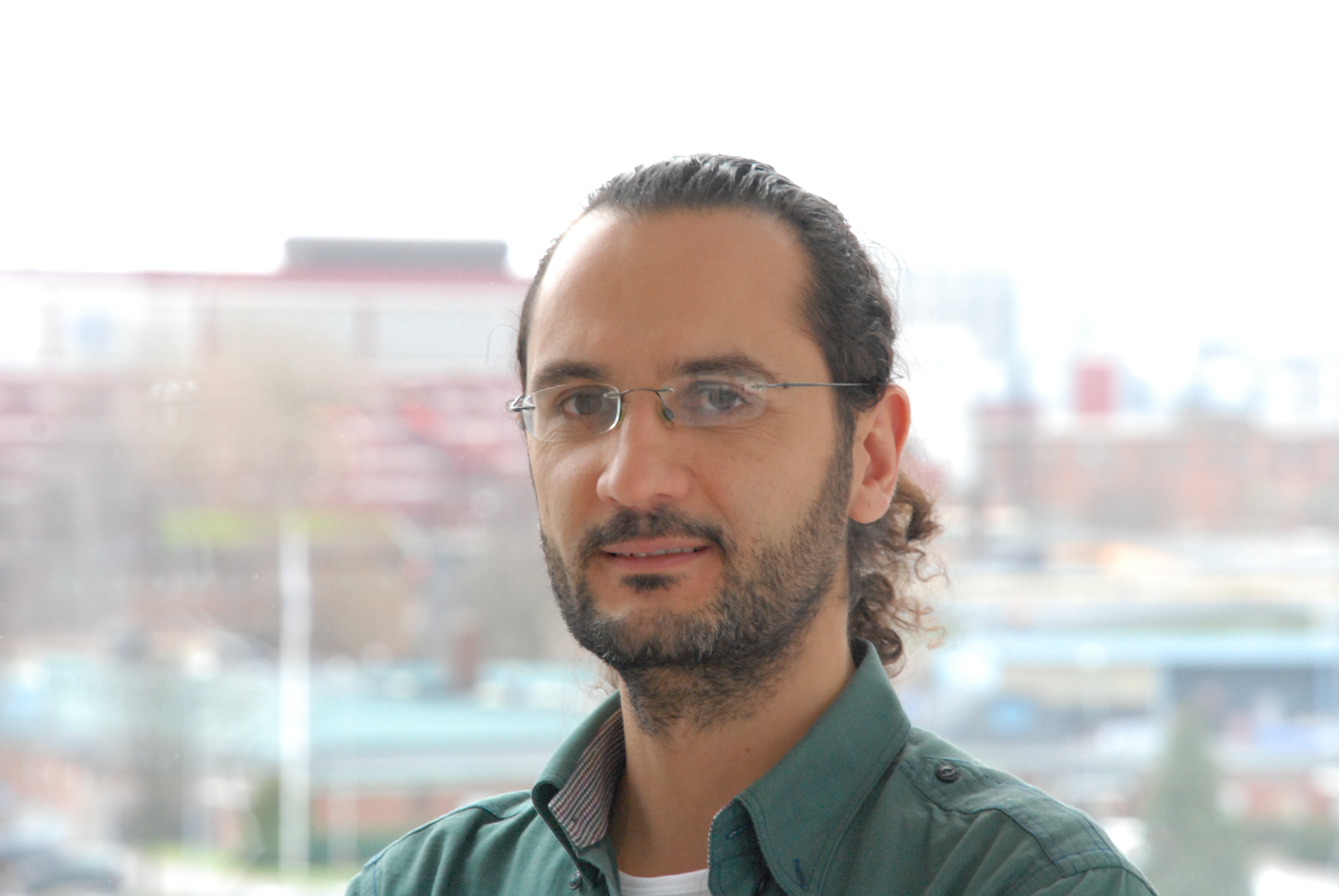}}]{Eren Erdal Aksoy}
received his M.Sc. degree in Mechatronics from the University of Siegen in Germany in 2008. He earned his Ph.D. degree in computer science from the University of Göttingen in Germany in 2012. He is currently employed as an Associate Professor at Halmstad University in Sweden. His research interests include semantic scene perception, computer vision, and robotics.
\end{IEEEbiography}

\begin{IEEEbiography}[{\includegraphics[width=1in,height=1.25in,clip,keepaspectratio]{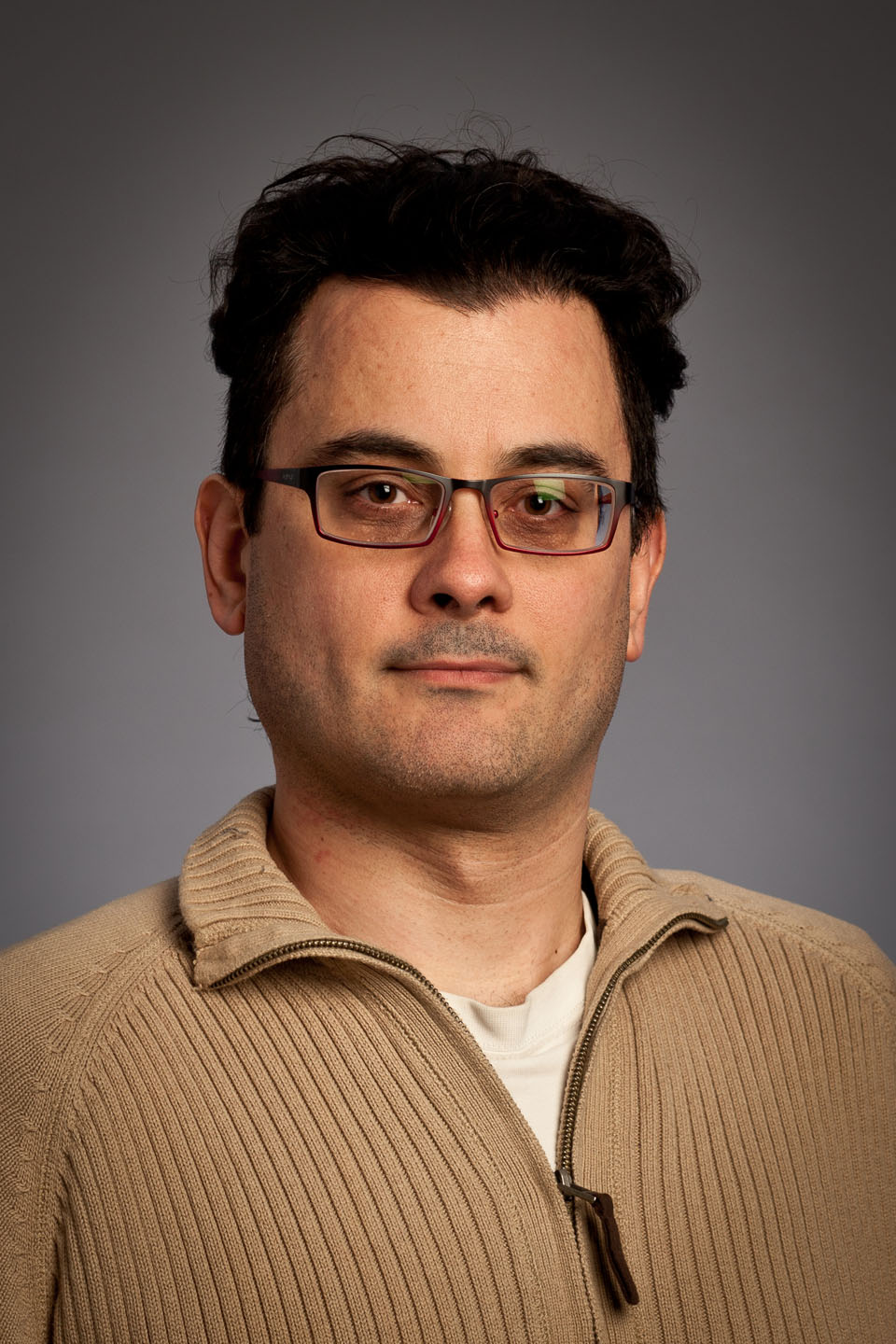}}]{Fernando Alonso-Fernandez}
is a docent and a Professor at Halmstad University, Sweden. He received the M.S./Ph.D. degrees in telecommunications from Universidad Politecnica de Madrid, Spain, in 2003/2008. Since 2010, he has been with Halmstad University, Sweden. His research interests include biometrics and computer vision for security applications.
\end{IEEEbiography}





\end{document}

%% file: preamble.tex
\newcommand{\OURSp}[1]{HYDRO{#1}} 
\newcommand{\OURS}[1]{{HYDRO}{#1}} 
\newcommand{\ESD}[1]{{ESD}{#1}} 

%% file: supplementary_material.tex
\clearpage

\newcounter{alphasect}
\def\alphainsection{0}

\let\oldsection=\section
\def\section{%
  \ifnum\alphainsection=1%
    \addtocounter{alphasect}{1}
  \fi%
\oldsection}%

\renewcommand\thesection{%
  \ifnum\alphainsection=1%
    \Alph{alphasect}
  \else%
    \arabic{section}
  \fi%
}%

\newenvironment{alphasection}{%
  \ifnum\alphainsection=1%
    \errhelp={Let other blocks end at the beginning of the next block.}
    \errmessage{Nested Alpha section not allowed}
  \fi%
  \setcounter{alphasect}{0}
  \def\alphainsection{1}
}{%
  \setcounter{alphasect}{0}
  \def\alphainsection{0}
}%

\begin{alphasection}

\begin{center}
  {\Large\bfseries Supplementary Material}\\[0.4em]
  {\large HYDRO: Towards Non-Reversible Face De-Identification Using a High-Fidelity Hybrid
Diffusion and Target-Oriented Approach}
\end{center}
\vspace{1em}
\textbf{Abstract.}~In the main manuscript, we provided a comprehensive evaluation of HYDRO on three different datasets in terms of: $(i)$ de-identification performance, $(ii)$ image fidelity, $(iii)$ attribute retention, $(iv)$ computational complexity, $(v)$ resilience to reconstruction attacks, $(vi)$ ablation studies, and $(vii)$ qualitative comparisons. In this supplementary material, we provide: $(i)$  additional information on two discriminators employed in HYDRO, $(ii)$ in-depth discussions on the adopted diffusion model and its training, $(iii)$ details on the reconstruction-attack model,  $(iv)$ formal definitions of the loss functions related to the attribute-preservation losses employed in the target-oriented generator of HYDRO, $(v)$ information on the reproducibility of HYDRO, $(vi)$ comparative results with video sequences, $(vii)$ ablation studies focused on the loss functions used with HYDRO, $(viii)$ comparisons of HYDRO and state-of-the-art models in terms of computational complexity and inference speed, $(ix)$ additional qualitative results that encompass further comparisons with more related works, along with analyses of edge and failure cases, and $(x)$ qualitative analyses of the attribute retention capabilities of HYDRO beyond the attributes explored in the main paper.

\begin{figure}[!b]
\centering
\includegraphics[width=0.45\textwidth]{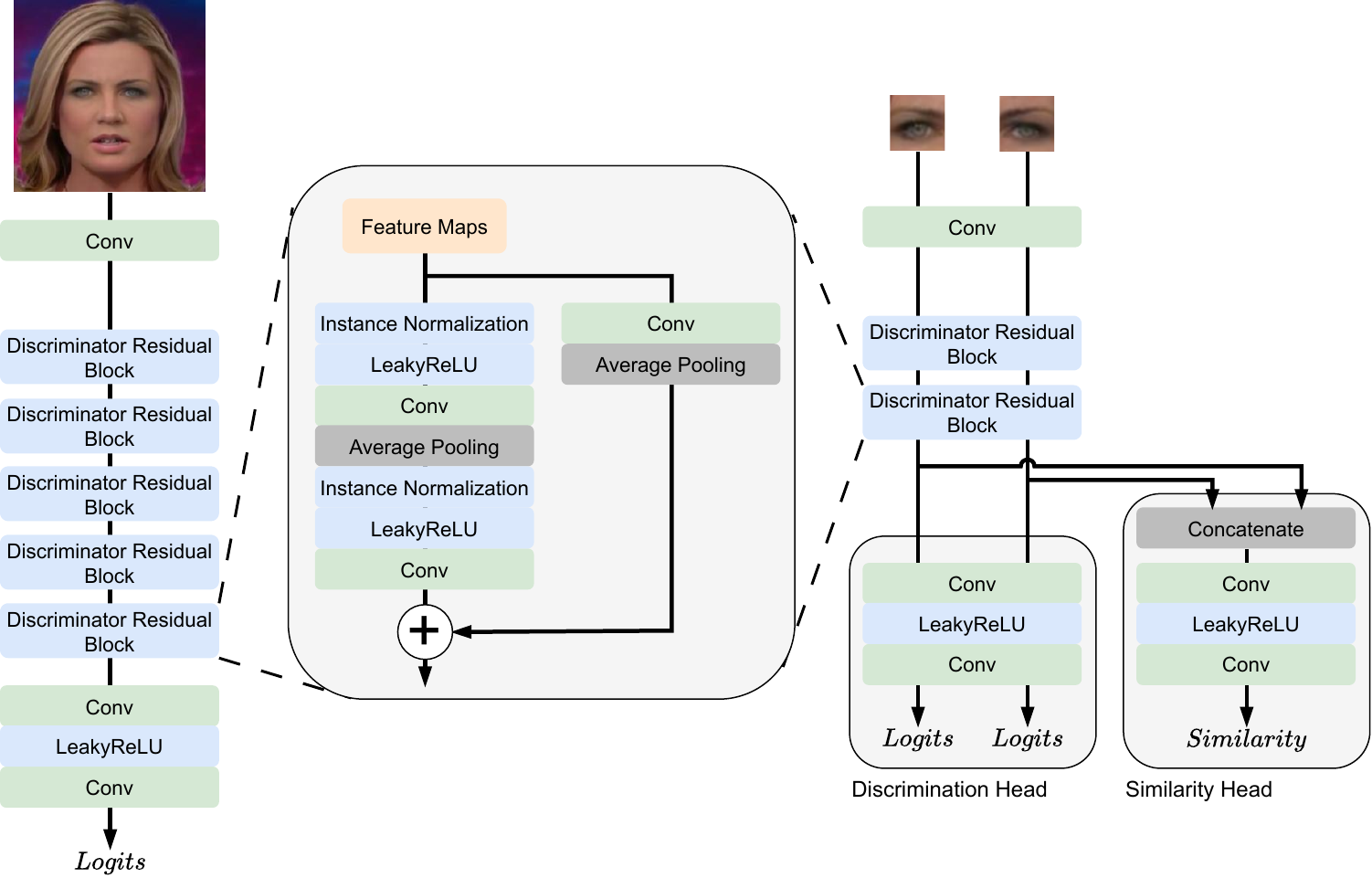}
\caption{Overview of the network architecture of the global discriminator (left) and the \emph{Eye Similarity Discriminator} (right). 
}
\label{fig:discriminators}
\end{figure}

\section{Discriminators}
\label{sec:sm:discrimnators}

In this section, we describe our two discriminators used to train \OURS~in greater detail. Fig. \ref{fig:discriminators} illustrates the architectural design of the global discriminator and the proposed {Eye Similarity Discriminator (ESD)}.

The two discriminators are trained in the same way, except for the use of the eye-similarity loss with ESD. We utilize the hinge loss for adversarial training. Let $x$ and $x_d$ represent the target image and de-identified image, respectively.~Then the hinge loss is:

\begin{equation}
    \mathcal{L}^D_{adv} = min(-D_k(x), 0) + min(D_k(x_d), 0),
\end{equation}

where $k\in\{glb, eye\}$ denotes either the global discriminator or the ESD. For the generator, the hinge loss is:

\begin{equation}
    \mathcal{L}^G_{adv} = min(-D_k(x_d), 0).
\end{equation}

Furthermore, we apply a gradient penalty \cite{wgangp} regularization term to stabilize the adversarial training as:

\begin{equation}
    \mathcal{L}^D_{gp} = \|\nabla_{x}D_k(x)\|_{L2}.
\end{equation}

where once again $k\in\{glb, eye\}$ denotes either the global discriminator or the ESD.

\section{Diffusion Model}
\label{sec:sm:diffusion}

\textbf{Diffusion Model Architecture:}~In this section, we provide a detailed overview of the diffusion model $\epsilon_{\theta}$ of HYDRO that ensures high-fidelity de-identification results. 
Fig.~\ref{fig:sm:diffusion} depicts the architecture of the diffusion model, which is similar to the de-identification generator $G$, presented in Fig.~1

in the main manuscript. It consists of a U-Net network with residual and attention blocks. The main difference compared to $G$ is in the application of a Multi-Layer Perceptron (MLP) for the identity vector, which is introduced to further process the identity embedding before feeding it to the intermediate blocks of the U-Net, as well as in the time step embedding, required for the diffusion process.

\begin{figure}[!t]
\centering
\includegraphics[width=0.45\textwidth]{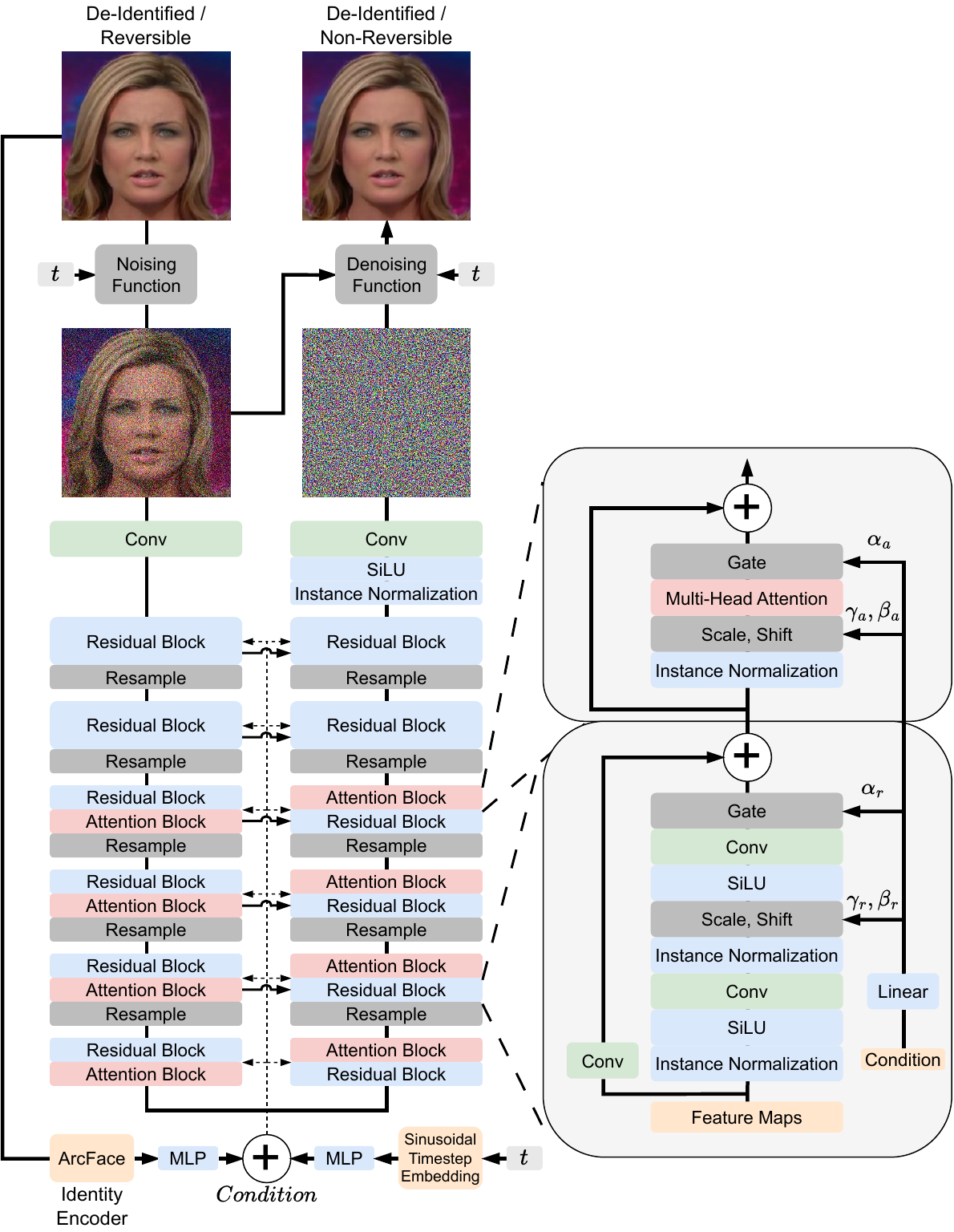}
\caption{Overview of the pixel-space diffusion model used to ensure robustness against reconstruction attack models.
}
\label{fig:sm:diffusion}
\end{figure}

\textbf{Diffusion Model Training:}~The diffusion model $\epsilon_{\theta}$ is trained using the denoising diffusion probabilistic model (DDPM) schedule \cite{DDPM}.~The diffusion model is directly trained in the pixel space.~The diffusion model is a $\epsilon$ prediction model.~Sampling is conducted using the DPM-Solver++ \cite{DPMSolver++} for one step with a low-strength noise in the final pipeline and this setting is used in all experiments.

For the forward diffusion, given a linear $\beta_t$ schedule, the schedule is defined as $\alpha_t=1-\beta_t$ for DDPM \cite{DDPM} and $\overline{\alpha}_{t}=\prod_{i=0}^t(1-\beta_i)$. We add noise to the sample $x_0$ through $x_t=\sqrt{\overline{\alpha}_{t}}x_0 + \sqrt{1 - \overline{\alpha}_{t}}\epsilon_0$, where $\epsilon_0\sim\mathcal{N}(0, I)$ represents noise sampled from a Gaussian distribution. We learn the reverse diffusion process by modeling the noise present in the sample $x_0$. The loss function is then:

\begin{equation}
    \mathcal{L}_{DDPM} = \|\epsilon - \epsilon_{\theta}(\sqrt{\overline{\alpha}_{t}}x_0 + \sqrt{1 - \overline{\alpha}}\epsilon_0)\|^2,
\end{equation}

where, as discussed in the main manuscript, $\epsilon_{\theta}$ is the diffusion model parameterized by $\theta$. Implementation details for the training and the DDPM schedule can be found in \secref{sec:sm:implementation}.

\section{Reconstruction-Attack Model}
\label{sec:sm:attack_model}

\begin{figure}[!t]
\centering
\includegraphics[width=0.45\textwidth, trim = 0 0 0 0mm, clip]{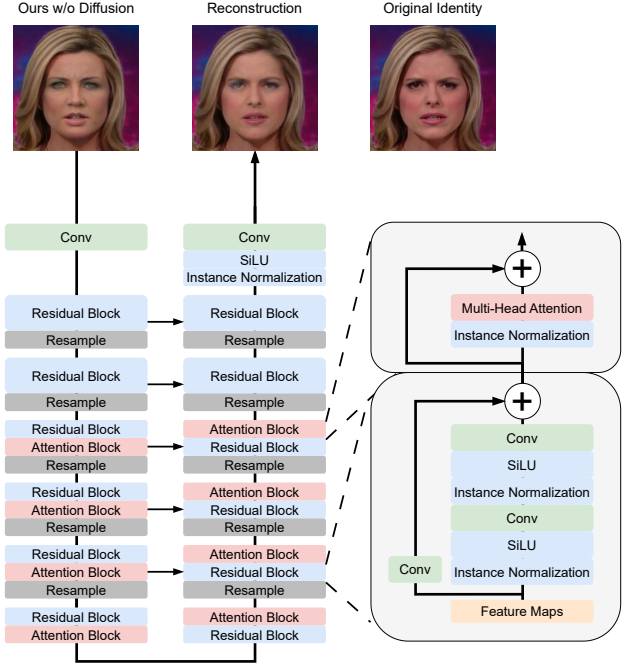}
\caption{Overview of the reconstruction-attack model $R$. In this particular case, we illustrate HYDRO \emph{without} the diffusion process, which results in a reconstruction that causes a match using face recognition models.
}
\label{fig:sm:reconstructor}
\end{figure}

A key component of the evaluation of HYDRO is the Reconstruction-Attack Model $R$ that is used to probe the robustness of HYDRO and other competing de-identification models to reconstruction attempts.~Below, we discuss the architectural design of the model, as well as  the training procedure used to learn its parameters.  

\textbf{Reconstruction Attack Model Architecture:} The reconstruction attack model $R$ is based on the U-Net model and shares the majority of its architectural details with the generator $G$ and diffusion model $\epsilon_{\theta}$ as shown in Fig. \ref{fig:sm:reconstructor}. However, there are two key differences: $(i)$ first, the reconstruction-attack model is an unconditional U-Net and, hence, is not conditioned on any information (e.g., identity, time-steps, etc.) and $(ii)$ second, it contains a single head that is responsible for predicting the initial target image before de-identification that still contains traces of the original identity and is, therefore, susceptible to reconstruction attacks. Given that a single head is used, the model obviously also does not return a mask.

Let the initial target face image be $x$ and let the corresponding de-identification version be denoted as $x_{d}$. The goal of $R$ is then to learn a mapping that can recover an approximation of the target face image, i.e.:

\begin{equation}
    R: x_{d} \mapsto \hat{x},
\end{equation}

where $\hat{x}$ is expected to contain the same identity information as $x$ in the case of a successful reconstruction attack.

\textbf{Reconstruction Attack Model Training:} The reconstruction attack model is trained over pairs of target face images and the respective de-identified counterparts. The training procedure assumes a \textit{black-box setting} simulating a bad-actor getting access to a sizable number of input and output images.  
One reconstruction model is trained for each state-of-the-art model that we compared with in the experimental evaluation conducted in the main manuscript. When learning the attack model for HYDRO, we utilize de-identified samples, on which the diffusion step was already applied. This setting is based on the assumption that if there is any leaked identity information, the reconstruction attack model will reliably learn to recover the original target identity.

Each reconstruction-attack model is trained using a combination of three losses, i.e., an L1 reconstruction loss, a perceptual loss \cite{LPIPS}, and an identity loss, similar to the one in the main manuscript, Eq.~8. The role of each loss is defined as follows:

\begin{itemize}
\item The L1 \textbf{reconstruction loss} $\mathcal{L}^R_{rec}$ encourages the model to produce outputs that share low-level pixel similarities with the actual input target face images, i.e.:  
\begin{equation}
    \mathcal{L}^R_{rec} = \|R(x_{d}) - x\|_{L1},
\end{equation}

where $\|\cdot\|_{L1}$ is the L1 norm.

\item The \textbf{perceptual loss} $\mathcal{L}_{lpips}$ is the Learned Perceptual Image Patch Similarity (LPIPS), described in \cite{LPIPS}, and ensures that higher-level perceptual characteristics of the recovered image $R(x_{d})$ are as close as possible to the original target face image $x$. The loss penalizes dissimilarities in feature maps computed from a VGG16 model pretrained on ImageNet \cite{imagenet}. We do not use the linear layers per feature map scale and directly calculate the loss between the feature maps.

\item The \textbf{identity loss} $\mathcal{L}^R_{id}$ is defined in a similar way as in the main manuscript but aims to minimize the difference in the identity information encoded through the identity encoder $I$, i.e.:

\begin{equation}
  \mathcal{L}^R_{id} = 1 - cos(I(x), I(R(x_d))),
  \label{eq:sm:identity_lossR}
\end{equation}
where $cos(\cdot)$ is the cosine similarity.
\end{itemize}

The overall learning objective of $R$ that is minimized during training is then defined as:

\begin{equation}
  \mathcal{L}_{R} = \lambda^R_{id}\mathcal{L}^R_{id} +\lambda^R_{rec}\mathcal{L}^R_{rec} + \mathcal{L}_{lpips},
  \label{eq:sm:overall}
\end{equation}
where $\lambda^R_{id}=5$ and $\lambda^R_{rec}=5$.

\section{Generator losses}
\label{s:sm:g_losses}
In the main manuscript, we describe the generator losses used to train the target-oriented de-identification model $G$. Here, we provide more formal definitions of the components of the attribute-preservation loss, which were omitted in the main manuscript due to space constraints. 

The \textbf{attribute-preservation loss} $\mathcal{L}_{att}^G$ aims to encourage the generator $G$ to retain as much of the attribute information from the input image $x$ in the de-identified result $x_d$ as possible. This is achieved through a combination of the following loss functions:
\begin{itemize}
    \item An L1 \textbf{reconstruction loss} $\mathcal{L}_{rec}^G$ between the input image $x$ and the de-identified output $x_d$ that penalizes pixel-level differences between the two images, i.e.,
    \begin{equation}
        \mathcal{L}_{rec}^G = \|G(x)-x\|_{L1},
    \end{equation}
    where $\|\cdot\|_{L1}$ is the L1 norm.
    \item A \textbf{feature-similarity loss} $\mathcal{L}_{fsr}^G$ between the intermediate feature representations of $x$ and $x_d$, extracted by the identity encoder ArcFace model \cite{FIVA, FaceDancer} that aims to minimize higher-level perceptual differences between the two images. Let $\phi_i$ denote the mapping function implemented by the ArcFace model at the $i$-th layer, the feature-similarity loss is then defined as
    \begin{equation}
        \mathcal{L}_{fsr}^G = \sum_{i\in\mathcal{A}}min(1 - cos(\phi_i(x), \phi_i(x_d)) - m_is, 0),
    \end{equation}
    where $\mathcal{A}$ is the set of considered model layers, and cos(·) is the cosine similarity, $m_i$ is a margin for the $i$-th layer, and $s$ is a hyperparameter for scaling $m$. We follow the settings in \cite{FIVA} and set all $m_i=0$. The layers used are the final feature maps for each resolution scale within the IResNet100 backbone \cite{IResNet}, one before each downsampling and one before the final batch normalization + flatten operation, resulting in a final number of layers of five.
    \item An L1 \textbf{distance loss} $\mathcal{L}_{shp}^G$ between the expression and pose coefficients, extracted through the 3DMM encoder from \cite{3DMMEncoder} that encourages pose and expression retention in the de-identified images. Let $c_{pose}^x$ and $c_{pose}^{x_d}$ denote the pose coefficients of the input target image $x$ and its de-identified version $x_d$, respectively. Likewise, let $c_{exp}^x$ and $c_{exp}^{x_d}$ stand for the respective expression coefficients. The distance loss $\mathcal{L}_{shp}^G$ is then defined as:
    \begin{equation}
        \mathcal{L}_{shp}^G = \|c_{pose}^x-c_{pose}^{x_d}\|_{L1}+\|c_{exp}^{x}-c_{exp}^{x_d}\|_{L1}.
    \end{equation}
    \item An L1 \textbf{distance gaze loss} $\mathcal{L}_{gze}^G$ between left- and right-eye heatmaps extracted by the heatmap-based landmark encoder from \cite{AdaptiveWing} that contributes towards better gaze preservation. Let the left- and right-eye patches be denoted as $x^l$ and $x^r$, respectively, and let $x^l_d$ and $x^r_d$ stand for their de-identified versions. If we denote the heatmap-generation function implemented within the landmarking model as $\psi$, then   the gaze loss can be defined as follows:
    \begin{equation}
        \mathcal{L}_{gze}^G = \sum_{i\in\{l,r\}} \|\psi(x^i)-\psi(x^i_d)\|_{L1}.
    \end{equation}
    \item An L1 \textbf{regularization term} $\mathcal{L}_{msk}^G$ over the blending mask $m$ to regularize the magnitude of the mask complexity, i.e.:
    \begin{equation}
        \mathcal{L}_{msk}^G = \|m\|_{L1}.
    \end{equation}
\end{itemize}

The complete attribute loss $\mathcal{L}_{att}^G$ is then defined as:
\begin{equation}
\resizebox{.87\hsize}{!}{$
  \mathcal{L}_{att}^G = \mathcal{L}_{fsr}^G + \lambda_{rec}^G\mathcal{L}_{rec}^G + \lambda_{msk}^G\mathcal{L}_{msk}^G \\
  + \lambda_{shp}^G\mathcal{L}_{shp}^G + \lambda_{gze}^G\mathcal{L}_{gze}^G ~,$
  \label{eq:sm:attribute_loss}
}
\end{equation}

with the balancing weights being $\lambda_{rec}^G=0.5$, $\lambda_{msk}^G=0.003$, $\lambda_{shp}^G=1$, and $\lambda_{gze}^G=25$, as introduced in the main manuscript. 

\section{Reproducibility}
\label{s:sm:reproducibility}
\subsection{Implementation Details}
\label{sec:sm:implementation}
The complete \OURS~model and all ablated variants are trained with the same configuration. We use AdamW \cite{AdamW} with a learning rate of $5e^{-5}$, $\beta_0 = 0.5$, $\beta_1 = 0.99$, weight decay of $2e^{-2}$, and epsilon value $\epsilon = 1e^{-8}$. The batch size is 10. Gradients are clipped at a max norm of 1 for all networks. The models are trained in mixed precision, autocasting to float16. We use a set seed (42 in our case) for all relevant packages used (e.g. PyTorch, Numpy). The models are learned in a $256\times256$ resolution and for $50,000$ iterations. The VGGFace2 \cite{vggface2} dataset is used for training, and we randomly load images for each batch.

When training the diffusion model $\epsilon_\theta$, we use a similar setting as with the training of \OURS. Specifically, we adopt AdamW \cite{AdamW} with a learning rate of $1e^{-4}$, $\beta^{opt}_0 = 0.95$, $\beta^{opt}_1 = 0.999$, weight decay of $1e^{-6}$, and epsilon value $\epsilon = 1e^{-8}$. The batch size is 10. Gradients are clipped at a max norm of 1 for all networks. The diffusion model $\epsilon_\theta$ is trained in mixed precision, autocasting to float16. We use a set seed (42 in our case) for all relevant packages used (e.g., PyTorch, Numpy). The diffusion model $\epsilon_\theta$ is trained in a $256\times256$ resolution. The diffusion model $\epsilon_\theta$ is trained for 150000 iterations. The VGGFace2 \cite{vggface2} dataset is used for training, and we randomly load images for each batch. The DDPM schedule \cite{DDPM} for applying noise to the images during training are used with the standard $\tau=1000$ time steps. $\beta_t$ starts at $0.00085$ ($\beta_0$) and ends at $0.012$ ($\beta_{\tau}$), with a linear schedule and leading time step spacing \cite{DDPMLeadingTimesteps}. The diffusion model $\epsilon_\theta$ is of noise $\epsilon$ prediction type.

\subsection{Identity Retrieval Details}
\label{ss:sm:id_details}

\begin{table}[!t]
\caption{Detailed information about the identity retrieval evaluation process. Shown are the used threshold for each FAR value together with the backbone used for each face recognition model. In this table $cos_d$ denotes cosine distance and $l2$ is the L2 distance.}
\begin{center}
\resizebox{\columnwidth}{!}{
\begin{tabular}{l|ccccc}
\toprule
Model       & Backbone          & Dataset                   & FAR = $10^{-3}$   & FAR = $10^{-4}$   & FAR = $10^{-5}$   \\
\midrule
CosFace     & IResNet50         & Glint360k~\cite{Glint360K} & $cos_d < 0.78$      & $cos_d < 0.72$      & $cos_d < 0.60$      \\
ArcFace     & IResNet100        & MS1MV3~\cite{MS1M}         & $cos_d < 0.76$      & $cos_d < 0.69$      & $cos_d < 0.62$      \\
FaceNet     & InceptionResNet   & VGGFace2~\cite{vggface2}   & $l2 < 0.93$         & $l2 < 0.71$         & $l2 < 0.59$      \\
AdaFace     & IResNet50         & MS1MV2~\cite{MS1M}         & $cos_d < 0.80$      & $cos_d < 0.75$      & $cos_d < 0.70$      \\
ElasticFace & IResNet100        & MS1MV2~\cite{MS1M}         & $cos_d < 0.80$      & $cos_d < 0.76$      & $cos_d < 0.70$      \\
\bottomrule

\end{tabular}
}
\end{center}
\label{t:id_details}
\end{table}

To facilitate reproducibility and fair comparisons, the identity retrieval evaluation is detailed in this section. Table \ref{t:id_details} shows the thresholds for each FAR value, backbone, and dataset used in the evaluation. The thresholds are derived by randomly sampling genuine and imposter pairs from the VGGFace2's train set \cite{vggface2}. 
To ensure reproducibility, we provide links to each evaluation model in the following footnote\footnote{\href{https://github.com/deepinsight/insightface/tree/master/recognition/arcface_torch}{CosFace}, \href{https://github.com/deepinsight/insightface/tree/master/recognition/arcface_torch}{ArcFace}, \href{https://github.com/timesler/facenet-pytorch}{FaceNet}, \href{https://github.com/mk-minchul/AdaFace}{AdaFace}, \href{https://github.com/fdbtrs/ElasticFace}{ElasticFace}}.

\subsection{Fréchet Inception Distance Details}
\label{ss:sm:fid_details}

To further facilitate reproducibility and fair comparisons, the setup for calculating FID is detailed in this section. The results for FID will typically vary depending on how the evaluation is set up.
In our case, we generate images from \OURS, ablations, and previous work and save them to a hard drive. The real and fake images are then loaded, followed by FID calculation. 
Each FID score is computed from a $192\times192$ face region, cropped from a $256\times256$ image, and resized to match the input of the Inception model, i.e., $299\times299$. The $192\times192$ facial area corresponds to the same region as typically used with recent face recognition models \cite{arcface, cosface, SphereFace}, where alignment is performed based on the five-point facial mark-up. See Fig. \ref{fig:fid_details} for an illustration of the process.

\textbf{FID Bias:} It is widely recognized that the FID score is susceptible to bias, as evidenced by the findings reported in \cite{FIDBias1, FIDBias2}. In terms of target-oriented face de-identification, it is also noteworthy that a perfect reconstruction would result in a perfect FID score. Regarding quantitative analysis, the FID score can be used as a reliable indicator for de-identification purposes when combined with other metrics. Specifically, a low identity retrieval rate and a low FID score indicate that the target distribution has been accurately matched while still fulfilling the de-identification objective. In instances where the identity retrieval rate is low, and the resulting qualitative output lacks visual credibility with regard to de-identification, the reconstruction attack identity retrieval rate may be regarded as a quantitative means of assessing this. The aforementioned scenario represents a theoretical form of adversarial noise that could potentially deceive face recognition models. However, our findings indicate that such noise could be effectively removed and reconstructed. Consequently, a low FID score, a low identity retrieval rate, and a low reconstruction attack identity retrieval rate demonstrate a robust alignment with the target distribution while quantitatively achieving de-identification.

\begin{figure}[!b]
\centering
\includegraphics[width=0.99\columnwidth]{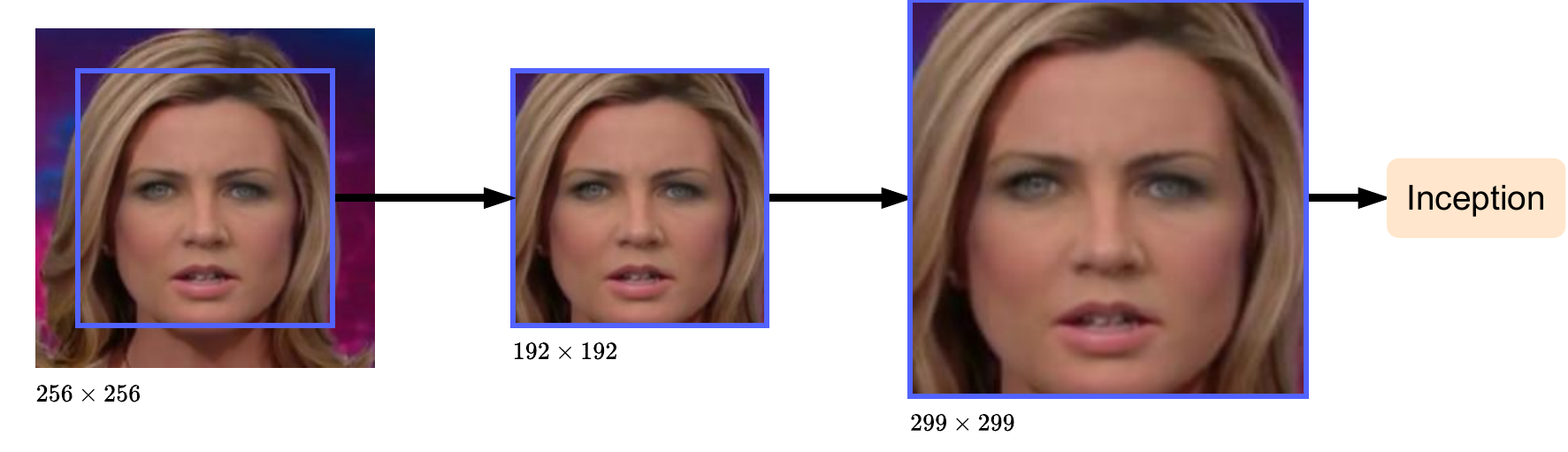}
\caption{Protocol used to evaluate and compare FID scores between \OURS, previous works, and ablations.}
\label{fig:fid_details}
\end{figure}

\section{Video Results}
Prior research has demonstrated the efficacy of target-oriented methods in processing video data, despite their frame-by-frame operational nature \cite{FIVA, FaceDancer, LIVEDEID, simswap, faceshifter}. Due to its target-oriented nature, HYDRO shares these characteristics and is capable to effectively de-identify video data, despite not incorporating an explicit component for ensuring temporal consistency or smoothness. 

To demonstrate~HYDRO's capabilities for video de-identification, we provide three example videos\footnote{\href{https://drive.google.com/drive/folders/1TRTZHxYZLaJRwTEb61k8zh-xlLC5qVzl?usp=sharing}{Video results}} as part of the uploaded supplementary material and compare our model qualitatively with FIVA \cite{FIVA} and G2Face \cite{G2Face} on video data. As can be seen from the videos, G2Face \cite{G2Face} in general produces convincing results, but exhibits inconsistencies between frames and fluctuations in the identity. A considerable issue are also side-profile faces, where the model struggles in producing realistic results (see first example video). FIVA produces comparably stronger results than G2FACE, but also exhibits difficulties with off-pose faces and face of smaller resolutions (as shown in the first example video). Due to the design of FIVA, artifacts are also visible in the chin area (see example videos two and three) and details around the eyes (see video two). The proposed HYDRO models, on the other hand, yields  convincing results even with faces in challenging poses and of poor resolution (see first example video), while working well with profile view faces (see example videos two and three). While some flickering can still be seen with the results produced by HYDRO, this is less evident than with the competing models and still very minute given that the videos we processed in a frame-by-frame manner.

\section{Inference}
\begin{table}[!b]
\caption{Quantitative inference cost comparisons, experiments ran on an Nvidia A6000. \textbf{Bold} indicates best, \underline{underline} indicates second best.}
\begin{center}
\resizebox{0.99\columnwidth}{!}{
\begin{tabular}{l|ccc}
\toprule
                                        & GFLOPS$\downarrow$  & Inference$\downarrow$ & Resolution \\
Model                                   &                   & [ms]              &            \\
\midrule
DeepPrivacy \cite{DEEPPRIVACY}          & 122.78            & \underline{15.51} & $256\times256$\\
CIAGAN \cite{CIAGAN}                    & \textbf{3.30}     & 29.83             & $128\times128$\\
RiDDLE \cite{RIDDLE}                    & 301.01           & 60.49             & $256\times256$\\ 
G2Face \cite{G2Face}                    & 280.02           & 43.98             & $256\times256$\\ 
FIVA \cite{FIVA}                        & 304.23            & 26.98             & $256\times256$\\
FIT  \cite{FIT}                         & \underline{30.09} & \textbf{12.00}    & $128\times128$\\
HYDRO                                   & 654.19            & 62.05             & $256\times256$\\
HYDRO $w \backslash o$ Diffusion        & 339.24            & 43.45             & $256\times256$\\
\hdashline
HYDRO $w \backslash o$ Identity Encoder & 629.98   & 37.5      & $256\times256$\\
HYDRO $w \backslash o$ Diffusion $w \backslash o$ Identity Encoder & 315.03   & 18.9      & $256\times256$\\
\bottomrule
\end{tabular}
}
\end{center}
\label{t:sm:inference}
\end{table}

Tab.~\ref{t:sm:inference} shows an analysis of the inference performance for a single batch of size $1$ on an Nvidia A6000 GPU for HYDRO and related works.~Naturally, adding a diffusion model increases the process time, as discussed in the main manuscript. DeepPrivacy \cite{DEEPPRIVACY} and FIT \cite{FIT} are the fastest methods.~However, note that DeepPrivacy is inpainting-based and has poor attribute-retention performance, while FIT operates on a lower resolution and, as a result, struggles with colorization artifacts. FIVA \cite{FIVA} is the fastest target-oriented method on a $256\times256$ resolution. HYDRO is almost $2.3$ times slower.
However, with the provided reconstruction defense and the fact that the diffusion step represent the most significant bottleneck in the inference speed, we see a fair trade-off for the privacy-protection level, resilience to reconstruction attacks, and computational complexity of HYDRO. The second computationally heavy component of HYDRO is the identity decoder, however, this encoder  offer significant potential for optimizations in the future, e.g., through knowledge distillation \cite{DISTILLATION_SURVEY}.

\end{alphasection}